\documentclass{article}

\usepackage[nonatbib, final]{neurips_2021}

\usepackage[utf8]{inputenc} 
\usepackage[T1]{fontenc}    
\usepackage[pagebackref=true,breaklinks=true,citecolor=blue,linkcolor=blue,colorlinks,urlcolor=blue,bookmarks=false]{hyperref}
\usepackage{url}            
\usepackage{booktabs}       
\usepackage{amsfonts}       
\usepackage{nicefrac}       
\usepackage{microtype}      
\usepackage{xcolor}         

\usepackage{amsmath} 
\usepackage{bm}
\usepackage{epsfig}
\usepackage{graphicx}
\usepackage{subcaption}
\usepackage{soul}
\usepackage{multirow}
\usepackage{pgffor}

\usepackage{booktabs}
\usepackage{bm}
\usepackage{algorithm, algpseudocode}
\usepackage{textcomp}
\usepackage{multirow}
\usepackage[outercaption]{sidecap}
\usepackage{makecell}
\usepackage{stmaryrd}
\usepackage{wrapfig}
\usepackage{caption}
\usepackage{enumitem}
\usepackage{subcaption}
\usepackage{pifont}  
\usepackage{array}
\usepackage{pifont}
\newcommand{\PreserveBackslash}[1]{\let\temp=\\#1\let\\=\temp}
\newcolumntype{C}[1]{>{\PreserveBackslash\centering}p{#1}}
\newcolumntype{R}[1]{>{\PreserveBackslash\raggedleft}p{#1}}
\newcolumntype{L}[1]{>{\PreserveBackslash\raggedright}p{#1}}
\newcommand{\cmark}{\ding{51}}%
\newcommand{\xmark}{\ding{55}}%

\newcommand{\itm}[1]{{\texttt{\small #1}}}

\newcommand{\etal}{\textit{et al}. }
\newcommand{\bx}{{\bm x}}

\newcommand{\by}{{y}}
\newcommand{\f}{{\mathbf{f}}}

\newcommand{\muz}[1]{{\textcolor{black}{#1}}}

\title{Intriguing Properties of Vision Transformers}

\author{%
  Muzammal Naseer$^{\dagger\star}$ \quad Kanchana Ranasinghe$^{+ \star}$ \quad Salman Khan$^{\star\dagger}$ \\ \textbf{Munawar Hayat$^{\mathparagraph}$ \quad Fahad Shahbaz Khan$^{\star\S}$ \quad Ming-Hsuan Yang$^{\ddagger\circ\nabla}$}\\
  $^{\dagger}$Australian National University, $^{\star}$Mohamed bin Zayed University of AI,  $^+$Stony Brook University, \\
  $^{\mathparagraph}$Monash University, $^{\S}$Link\"{o}ping University, $^{\ddagger}$University of California, Merced, \\
  $^{\circ}$Yonsei University, $^{\nabla}$Google Research\\
  \texttt{muzammal.naseer@anu.edu.au} 
}

\begin{document}

\maketitle

\begin{abstract}
Vision transformers (ViT) have demonstrated impressive performance across numerous machine vision tasks. 
These models are based on multi-head self-attention mechanisms that can flexibly attend to a sequence of image patches to encode contextual cues. 
%
An important question is how such  flexibility (in attending image-wide context conditioned on a given patch) can facilitate handling nuisances in natural images e.g., severe occlusions, domain shifts, spatial permutations, adversarial and natural perturbations. 
We systematically study this question via an extensive set of experiments encompassing three ViT families and provide comparisons with a high-performing convolutional neural network (CNN). 
We show and analyze the following intriguing properties of ViT: \itm{(a)} Transformers are highly robust to severe occlusions, perturbations and domain shifts, \emph{e.g.,} retain as high as 60\% top-1 accuracy on ImageNet even after randomly occluding 80\% of the image content. 
\itm{(b)} The robustness towards occlusions is not due to texture bias, instead we show that ViTs are significantly less biased towards local textures, compared to CNNs. 
When properly trained to encode shape-based features, ViTs demonstrate shape recognition capability comparable to that of human visual system, previously unmatched in the literature. 
\itm{(c)} Using ViTs to encode shape representation leads to an interesting consequence of accurate semantic segmentation without pixel-level supervision. 
\itm{(d)} Off-the-shelf features from a single ViT model can be combined to create a feature ensemble, leading to high accuracy rates across a range of classification datasets in both traditional and few-shot learning paradigms. 
We show effective features of ViTs are due to flexible and dynamic receptive fields possible via self-attention mechanisms. 
Code: \url{https://git.io/Js15X}.
\end{abstract}

\section{Introduction}
As visual transformers (ViT) attract more interest \cite{khan2021transformers}, it becomes highly pertinent to study characteristics of their learned representations. 
Specifically, from the perspective of safety-critical applications such as autonomous cars, robots and healthcare; the learned representations need to be robust and generalizable. 
In this paper, we compare the performance of transformers  with convolutional neural networks (CNNs) for handling nuisances (\emph{e.g.,} occlusions, distributional shifts, adversarial and natural perturbations) and generalization across different data distributions. 
Our in-depth analysis is based on \emph{three} transformer families, ViT \cite{dosovitskiy2020image}, DeiT \cite{touvron2020deit} and T2T \cite{yuan2021tokens} across \emph{fifteen} vision datasets. 
For brevity, we refer to all the transformer families as ViT, unless otherwise mentioned. 

We are intrigued by the fundamental differences in the operation of convolution and self-attention, that have not been extensively explored in the context of robustness and generalization. 
While convolutions excel at learning local interactions between elements in the input domain (\emph{e.g.,} edges and contour information), self-attention has been shown to effectively learn global interactions (\emph{e.g.,} relations between distant object parts) \cite{ramachandran2019stand,hu2019local}. 
Given a query embedding, self-attention finds its interactions with the other embeddings in the sequence, thereby conditioning on the local content while modeling global relationships \cite{vaswani2021scaling}. 
In contrast, convolutions are content-independent as the same filter weights are applied to all inputs regardless of their distinct nature. %
Given the content-dependent long-range interaction modeling capabilities, our analysis shows that ViTs can flexibly adjust their receptive field to cope with nuisances in data and enhance expressivity of the representations.  

\begin{figure}[t!]
    \centering
        \centering
        \includegraphics[width=\linewidth]{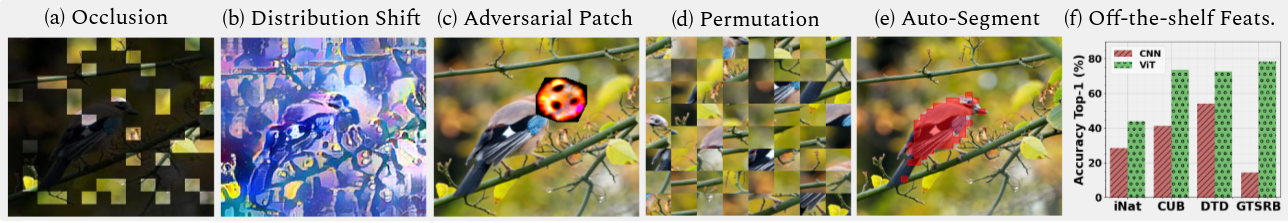}
        \vspace{-1.5em}
         \caption{We show intriguing properties of ViT including impressive robustness to  \itm{(a)} severe occlusions,  \itm{(b)} distributional shifts (\emph{e.g.,} stylization to remove texture cues), \itm{(c)} adversarial perturbations, and \itm{(d)} patch permutations. 
         Furthermore, our ViT models trained to focus on shape cues can segment foregrounds without any pixel-level supervision \itm{(e)}. 
         Finally, off-the-shelf features from ViT models generalize better than CNNs \itm{(f)}.}
    \label{fig:main_demo}
    \end{figure}
    
Our systematic experiments and novel design choices lead to the following interesting findings: 
\begin{itemize}[noitemsep,leftmargin=.75cm,topsep=0pt]
    \item ViTs demonstrate strong robustness against severe occlusions for foreground objects, non-salient background regions and random patch locations, when compared with state-of-the-art CNNs. 
    For instance, with a significant random occlusion of up to 80\%, DeiT \cite{touvron2020deit} can maintain top-1 accuracy up to $\sim$60\% where CNN has zero accuracy, on ImageNet \cite{ILSVRC15} val. set.
    \item When presented with texture and shape of the same object, CNN models often make decisions based on texture  \cite{geirhos2018imagenet}.
    In contrast, ViTs perform better than CNNs 
    and comparable to humans on shape recognition.
    This highlights robustness of ViTs to deal with significant distribution shifts \emph{e.g.,} recognizing object shapes in less textured data such as paintings. 
    \item Compared to CNNs, ViTs show better robustness against other nuisance factors such as spatial patch-level permutations, adversarial perturbations and common natural corruptions (\emph{e.g.,} noise, blur, contrast and pixelation artefacts).  
    However, similar to CNNs \cite{mummadi2021does}, a shape-focused training process renders them vulnerable against adversarial attacks and common corruptions.
    \item Apart from their promising robustness properties,  off-the-shelf ViT features from ImageNet pretrained models generalize exceptionally well to new domains \emph{e.g.,} few-shot learning, fine-grained recognition, scene categorization and long-tail classification settings. 
\end{itemize}

In addition to our extensive experimental analysis and new findings, we introduce several novel design choices to highlight the strong potential of ViTs. 
To this end, we propose an architectural modification to DeiT to encode shape-information via a dedicated token that demonstrates how seemingly contradictory cues can be modeled with distinct tokens within the same architecture, leading to favorable implications such as automated segmentation without pixel-level supervision.
Moreover, our off-the-shelf feature transfer approach utilizes an ensemble of representations derived from a single architecture to obtain state-of-the-art generalization with a pre-trained ViT (Fig.~\ref{fig:main_demo}).

\section{Related Work}
CNNs have shown state-of-the-art performance in independent and identically distributed (i.i.d) settings but remain highly sensitive to distributional shifts; adversarial noise \cite{szegedy2013intriguing, naseer2019cross}, common image corruptions \cite{hendrycks2019benchmarking}, and domain shifts (\emph{e.g.,} RGB to sketches) \cite{li2017deeper}.
It is natural to ask if ViT, that processes inputs based on self-attention, offers any advantages in comparison to CNN.
Shao \etal \cite{shao2021adversarial} analyze ViTs against adversarial noise and show ViTs are more robust to high frequency changes. 
Similarly, Bhojanapalli \etal \cite{bhojanapalli2021understanding} study ViT against spatial perturbations \cite{shao2021adversarial} and its robustness to removal of any single layer. 
Since ViT processes image patches, we focus on their robustness against  patch masking, localized adversarial patches \cite{brown2017adversarial} and common natural corruptions. 
A concurrent work from Paul and Chen \cite{paul2021vision} also develops similar insights on robustness of ViTs but with a somewhat different set of experiments.

Geirhos \etal \cite{geirhos2018imagenet} provide evidence that CNNs mainly exploit texture to make a decision and give less importance to global shape. 
This is further confirmed by CNN ability to only use local features \cite{brendel2019approximating}. 
Recently, \cite{islam2021shape} quantifies mutual information \cite{foster2011lower} between shape and texture features. 
Our analysis indicates that large ViT models have less texture bias and give relatively higher emphasis to shape information. 
ViT's shape-bias approaches human-level performance when directly trained on stylized ImageNet \cite{geirhos2018imagenet}. 
Our findings are consistent with a concurrent recent work that demonstrates the importance of this trend on human behavioural understanding and bridging the gap between human and machine vision \cite{tuli2021convolutional}.
A recent work \cite{caron2021emerging} shows that self-supervised ViT can automatically segment foreground objects. In comparison, we show how shape-focused learning can impart similar capability in the image-level supervised ViT models, without any pixel-level supervision. 

Zeiler \etal \cite{zeiler2014visualizing} introduce a method to visualize CNN features at different layers and study the performance of off-the-shelf features.
In a similar spirit, we study the generalization of off-the-shelf features of ViT in comparison to CNN. Receptive field is an indication of network's ability to model long range dependencies. 
The receptive field of Transformer based models covers the entire input space, a property that resembles handcrafted features \cite{yang2013saliency}, but ViTs have higher representative capacity. 
This allows ViT to model global context and preserve the structural information compared to CNN \cite{mao2021transformer}. 
This work is an effort to demonstrate the effectiveness of flexible receptive field and content-based context modeling in ViTs towards robustness and generalization of the learned features.

\section{Intriguing Properties of Vision Transformers}

\subsection{Are Vision Transformers Robust to Occlusions?}
\label{sec:patch_drop}
The receptive field of a ViT spans over the entire image and it models the interaction between the sequence of image patches using self-attention \cite{mao2021transformer,vaswani2017attention}. 
We study whether ViTs perform robustly in occluded scenarios, where some or most of the image content is missing. 

\textbf{Occlusion Modeling:} Consider a network $\f$, that processes an input image $\bx$ to predict a label $\by$, where $\bx$ is represented as a patch sequence with $N$ elements, \emph{i.e.,}  $\bx = \{x_{i}\}_{i=1}^N$ \cite{dosovitskiy2020image}. 
While there can be multiple ways to define occlusion, we adopt a simple masking strategy, where we select a subset of the total image patches, $M < N$, and set pixel values of these patches to zero to create an occluded image, $\bx'$. 
We refer to this approach as {PatchDrop}. The objective is then to observe robustness such that $\f(\bx')_{\text{argmax}}=\by$. We experiment with three variants of our occlusion approach, \textbf{(a)} Random PatchDrop, \textbf{(b)} Salient (foreground) PatchDrop, and \textbf{(c)} Non-salient (background) PatchDrop. 

\begin{figure}[b!]
    \centering
    \begin{minipage}{.38\textwidth}
    \caption{\small An example image with its occluded versions (Random, Salient and Non-Salient). The occluded images are correctly classified by Deit-S \cite{touvron2020deit} but mis-classified by ResNet50 \cite{he2016deep}. Pixel values in occluded (black) regions are set to zero.
    }
     \label{fig:patchdrop_vis}
    \end{minipage}
    \begin{minipage}{.6\textwidth}\vspace{-1em}
    \begin{subfigure}[b]{0.24\linewidth}        
        \centering
        \caption*{\tiny Original Image}
        \vspace{-0.5em}
        \includegraphics[width=\linewidth]{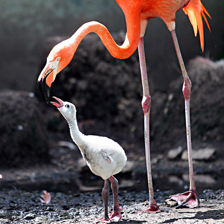}
    \end{subfigure}
        \begin{subfigure}[b]{0.24\linewidth}        
        \centering
        \caption*{\tiny Random PatchDrop}
        \vspace{-0.5em}
        \includegraphics[width=\linewidth]{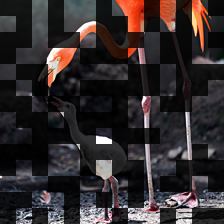}
    \end{subfigure}    
        \begin{subfigure}[b]{0.24\linewidth}        
        \centering
        \caption*{\tiny Salient PatchDrop}
        \vspace{-0.5em}
        \includegraphics[width=\linewidth]{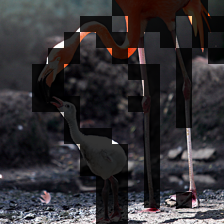}
    \end{subfigure}    
        \begin{subfigure}[b]{0.24\linewidth}        
        \centering
        \caption*{\tiny Non-Salient PatchDrop}
        \vspace{-0.5em}
        \includegraphics[width=\linewidth]{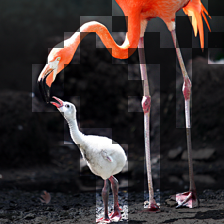}
    \end{subfigure}
    \end{minipage}
    \vspace{-1em}
\end{figure}

\textit{Random PatchDrop:} A subset of $M$ patches is randomly selected and dropped (Fig.~\ref{fig:patchdrop_vis}). 
Several recent Vision Transformers \cite{dosovitskiy2020image,touvron2020deit,yuan2021tokens} divide an image into 196 patches belonging to a 14x14 spatial grid; i.e. an image of size 224$\times$224$\times$3 is split into 196 patches, each of size 16$\times$16$\times$3. 
As an example, dropping 100 such patches from the input is equivalent to losing 51\% of the image content.

\textit{Salient (foreground) PatchDrop:} 
Not all pixels have the same importance for vision tasks. 
Thus, it is important to study the robustness of ViTs against occlusions of highly salient regions. 
We leverage a self-supervised ViT model DINO \cite{caron2021emerging} that is shown to effectively segment salient objects. 
In particular, the spatial positions of information flowing into the final feature vector (class token) within the last attention block are exploited to locate the salient pixels. 
This allows to control the amount of salient information captured within the selected pixels by thresholding the quantity of attention flow.

We select the subset of patches containing the top $Q$\% of foreground information (deterministic for fixed $Q$) and drop them. 
Note that this $Q$\% does not always correspond to the pixel percentage, \emph{e.g.,} 50\% of the foreground information of an image may be contained within only 10\% of its pixels.

\textit{Non-salient (background) PatchDrop:}
The least salient regions of the image are selected following the same approach as above, using \cite{caron2021emerging}. 
The patches containing the lowest $Q$\% of foreground information are selected and dropped here. 
Note this does not always correspond to the pixel percentage, \emph{e.g.,} 80\% of the pixels may only contain 20\% of the non-salient information for an image.

\begin{figure}[t!]
    \centering
    \begin{subfigure}[b]{0.24\linewidth}        
        \centering
        \includegraphics[width=\linewidth]{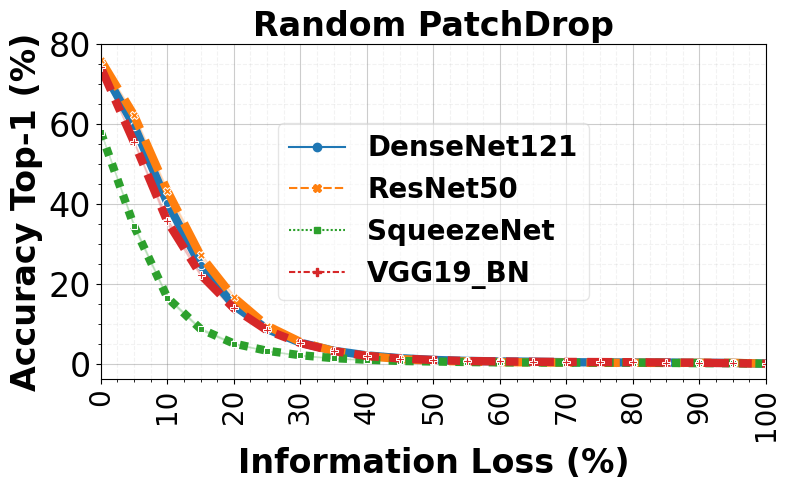}
    \end{subfigure}
    \begin{subfigure}[b]{0.24\linewidth}        
        \centering
        \includegraphics[width=\linewidth]{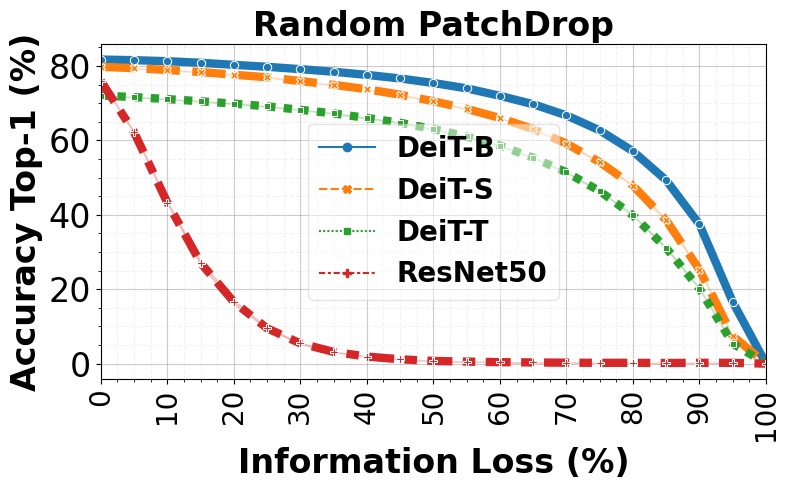}
    \end{subfigure}
    \begin{subfigure}[b]{0.24\linewidth}        
        \centering
        \includegraphics[width=\linewidth]{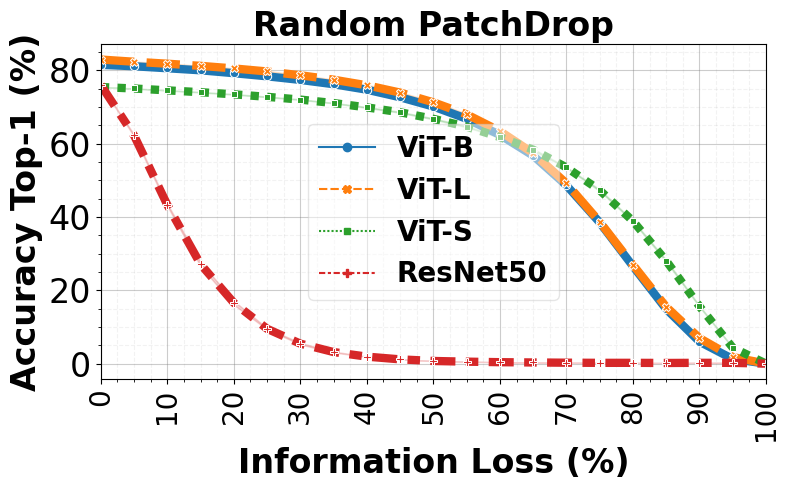}
    \end{subfigure}
    \begin{subfigure}[b]{0.24\linewidth}        
        \centering
        \includegraphics[width=\linewidth]{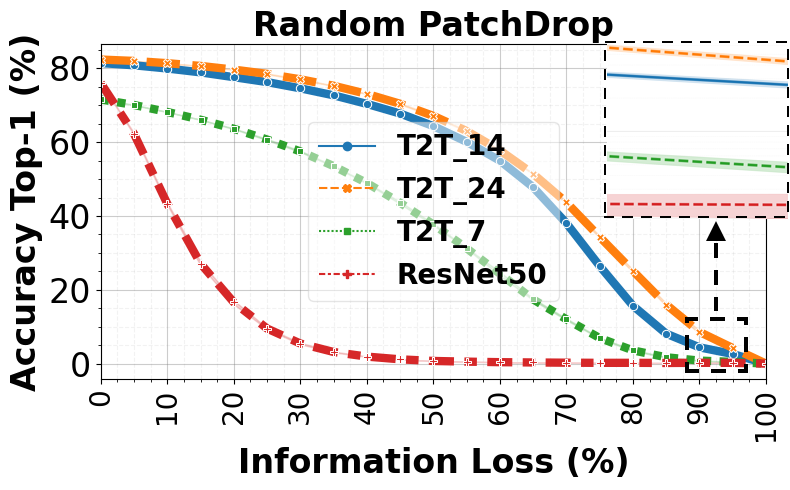}
    \end{subfigure}
    \vspace{0.5em}
    
    \begin{subfigure}[b]{0.24\linewidth}        
        \centering
        \includegraphics[width=\linewidth]{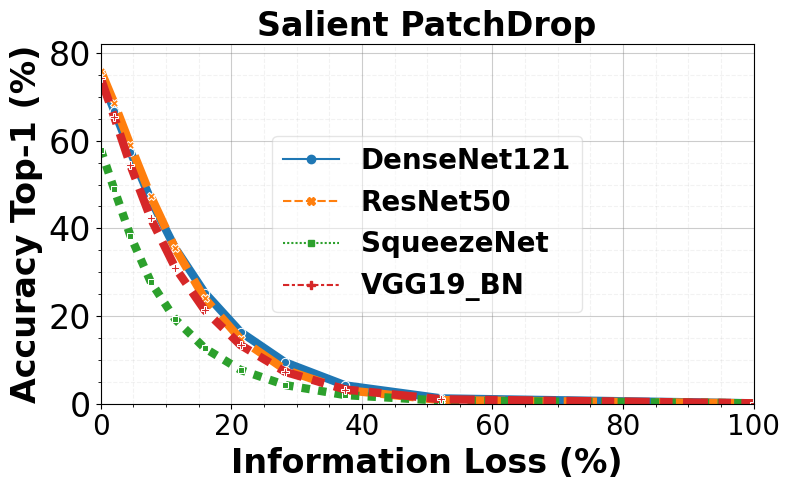}
    \end{subfigure}
    \begin{subfigure}[b]{0.24\linewidth}        
        \centering
        \includegraphics[width=\linewidth]{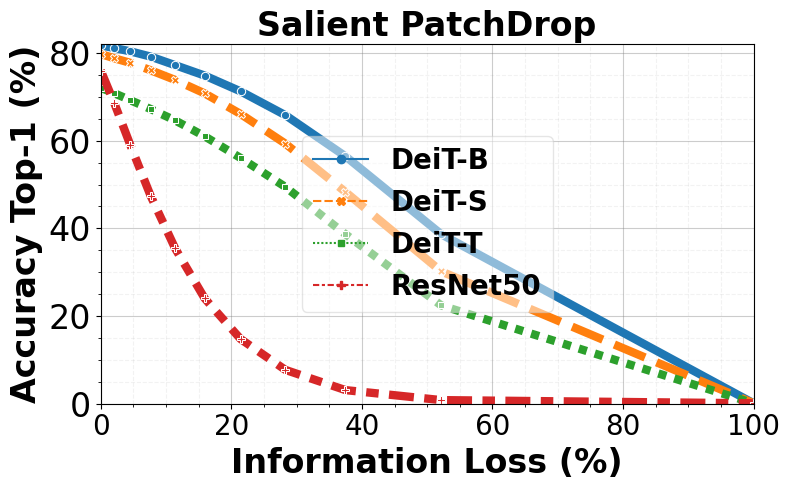}
    \end{subfigure}
    \begin{subfigure}[b]{0.24\linewidth}        
        \centering
        \includegraphics[width=\linewidth]{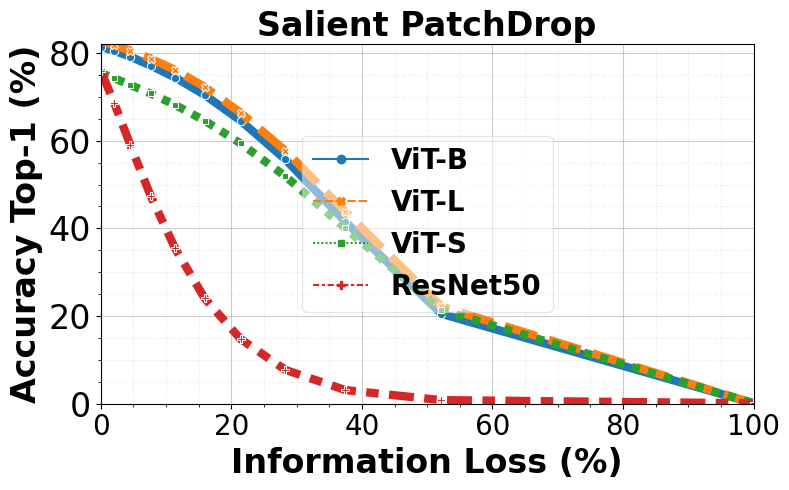}
    \end{subfigure}
    \begin{subfigure}[b]{0.24\linewidth}        
        \centering
        \includegraphics[width=\linewidth]{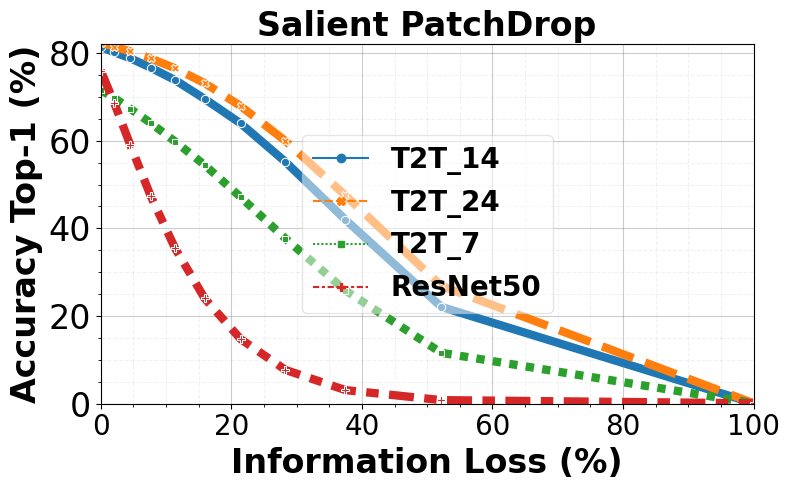}
    \end{subfigure}
    \vspace{0.5em}
    
    \begin{subfigure}[b]{0.24\linewidth}        
        \centering
        \includegraphics[width=\linewidth]{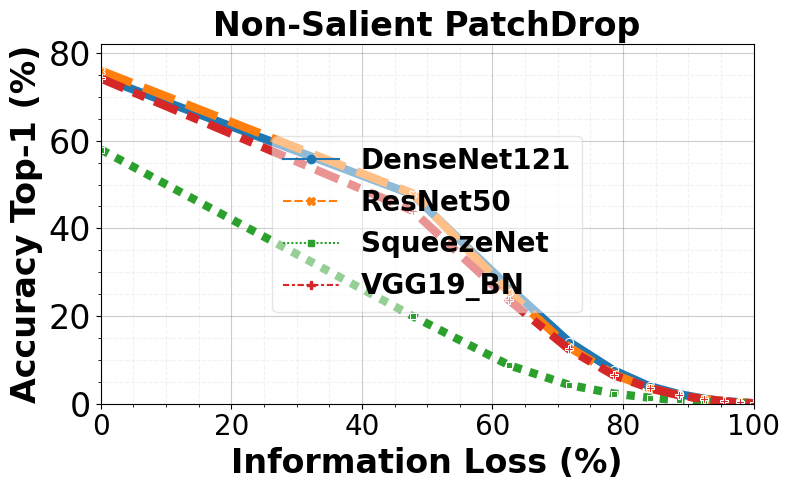}
    \end{subfigure}
    \begin{subfigure}[b]{0.24\linewidth}        
        \centering
        \includegraphics[width=\linewidth]{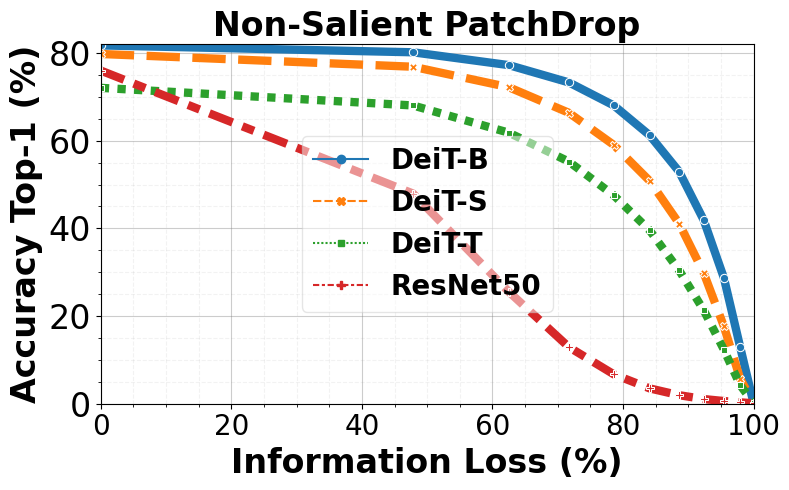}
    \end{subfigure}
    \begin{subfigure}[b]{0.24\linewidth}        
        \centering
        \includegraphics[width=\linewidth]{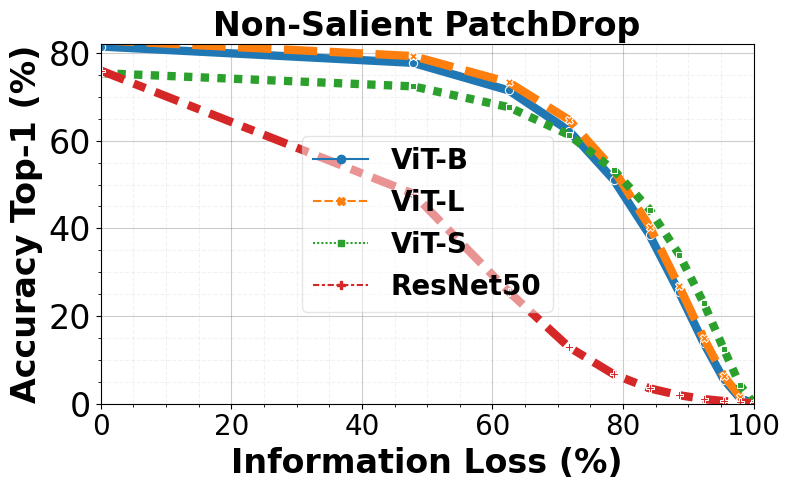}
    \end{subfigure}
    \begin{subfigure}[b]{0.24\linewidth}        
        \centering
        \includegraphics[width=\linewidth]{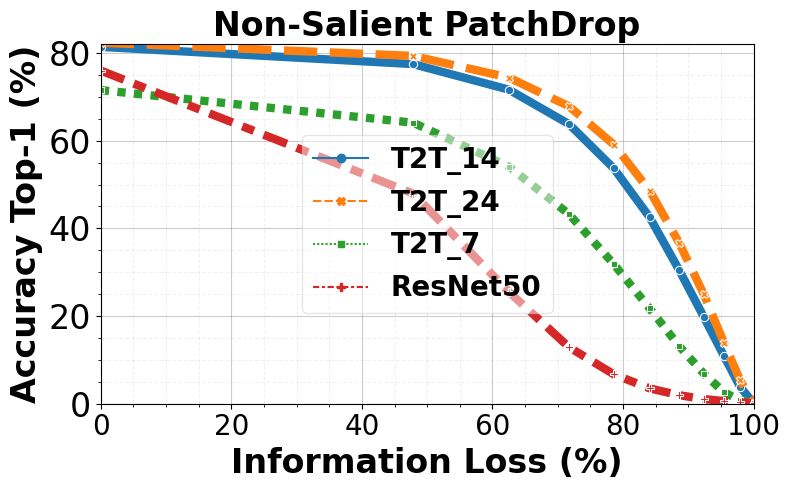}
    \end{subfigure}
    \vspace{0.5em}
    
     \caption{\small Robustness against object occlusion in images is studied under three PatchDrop settings (see Sec~\ref{sec:patch_drop}). (\emph{left}) We study the robustness of CNN models to occlusions, and identify ResNet50 as a strong baseline. (\emph{mid-left}) We compare the DeiT model family against ResNet50 exhibiting their superior robustness to object occlusion. (\emph{mid-right}) Comparison against ViT model family. (\emph{right}) Comparison against T2T model family.}
     \label{fig:patch_drop_plots}
     \vspace{-1.5em}
\end{figure}

\textbf{Robust Performance of Transformers Against Occlusions:}
We consider visual recognition task with models pretrained on ImageNet \cite{dosovitskiy2020image}.
The effect of occlusion is studied on the validation set (50k images). 
We define information loss (IL) as the ratio of dropped and total patches ($M$ / $N$). IL is varied to obtain a range of occlusion levels for each PatchDrop methodology. 
The results (Top-1 \%) reported in  Fig.~\ref{fig:patch_drop_plots} show significantly robust performance of ViT models against CNNs. 
In the case of random PatchDrop, we report the mean of accuracy across 5 runs. 
For Salient and Non-Salient Patchdrop, we report the accuracy values over a single run, since the occlusion mask is deterministic.
CNNs perform poorly when 50\% of image information is randomly dropped. 
For example, ResNet50 (23 Million parameters) achieves 0.1\% accuracy in comparison to DeiT-S (22 Million parameters) which obtains 70\% accuracy when 50\% of the image content is removed. 
An extreme example can be observed when 90\% of the image information is randomly masked but Deit-B still exhibits 37\% accuracy. 
This finding is consistent among different ViT architectures \cite{dosovitskiy2020image, touvron2020deit, yuan2021tokens}.
Similarly, ViTs show significant robustness to the foreground (salient) and background (non-salient) content removal. 
{See Appendix \ref{sec:different_patch_Size_with_random_PatchDrop}, \ref{sec:random_pixeldrop}, \ref{sec:feature_drop}, \ref{sec:occultions_swin_regnety}, and \ref{sec:shape_biased} for further results on robustness analysis.}

\begin{figure}[t]
    \centering
    \begin{subfigure}[b]{\linewidth}        
        \centering
        \includegraphics[width=\linewidth]{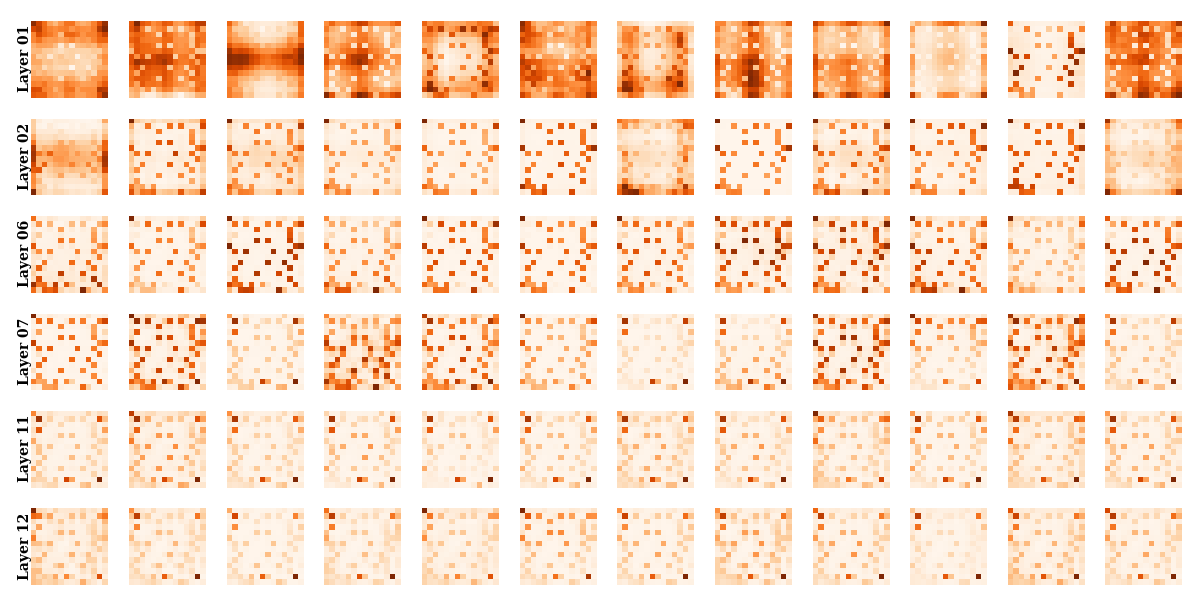}
    \end{subfigure}
    \begin{minipage}[t]{0.90\textwidth}
     \vspace{-1em}
     \caption{\small Attention maps (averaged over the entire ImageNet val. set) relevant to each head in multiple layers of an ImageNet pre-trained DeiT-B model. All images are occluded (Random PatchDrop) with the same mask (bottom right). Observe how later layers clearly attend to non-occluded regions of images to make a decision, an evidence of the model's highly dynamic receptive field.}
     \label{fig:patchdrop_attention_maps}
    \end{minipage} 
    \begin{minipage}[t]{0.07\textwidth}
    \strut\vspace*{-\baselineskip}\newline\includegraphics[width=\linewidth]{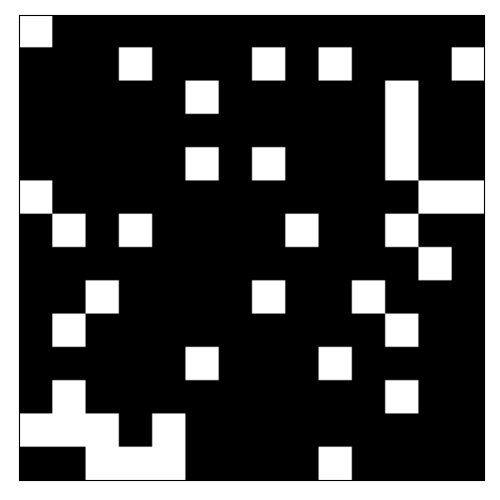}
    \end{minipage} 
\end{figure}

\muz{\textbf{ViT Representations are Robust against Information Loss:}} In order to better understand model behavior against such occlusions, we visualize the attention (Fig.~\ref{fig:patchdrop_attention_maps}) from each head of different layers. 
While initial layers attend to all areas, deeper layers tend to focus more on the leftover information in non-occluded regions of an image.  
We then study if such changes from initial to deeper layers lead to token invariance against occlusion which is important for classification. 
\muz{We measure the correlation coefficient between features/tokens of original and occluded images by using $corr(u, v) = \frac{ \sum_i \hat{u_i} \hat{v_i} } {n}$, where $\hat{u_i} = \frac {u_i - E[u_i]} {\sigma(u_i)}$, $E[\cdot]$ and $\sigma(\cdot)$ are mean and standard deviation operations \cite{forsyth2018probability}. In our case, random variables $u$ and $v$ refer to the feature maps for an original and occluded image defined over the entire ImageNet validation set.} 
In the case of ResNet50, we consider features before the logit layer and for ViT models, class tokens are extracted from the last transformer block. 
Class tokens from transformers are significantly more robust and do not suffer much information loss as compared to ResNet50 features (Table~\ref{tbl:correlation_coefficint_val_set}).
Furthermore, we visualize the correlation coefficient across the 12 selected superclasses within ImageNet hierarchy and note that the trend holds across different class types, even for relatively small object types such as \emph{insects}, \emph{food} items and \emph{birds} (Fig.~\ref{fig:correlation_coefficint_classwise_val_set}).
{See Appendix~\ref{sec:dynamic_receptive_field} for attention visualizations and~\ref{sec:supplementary_qualitative} for the qualitative results.}


Given the intriguing robustness of transformer models due to dynamic receptive fields and discriminability preserving behaviour of the learned tokens, an ensuing question is whether the learned representations in ViTs are biased towards texture or not.
One can expect a biased model focusing only on texture to still perform well when the spatial structure for an object is partially lost. 

\begin{table}[H]
\begin{minipage}{0.49\linewidth}
\vspace{-2em}
\centering
\small
\label{tab:correlation_coefficint_val_set}
\scalebox{0.85}{
\begin{tabular}{l|c|c|c}
\toprule
Model & \multicolumn{3}{c}{Correlation Coefficient: Random PatchDrop}\\
\cline{2-4}
         & 25\% Dropped   & 50\% Dropped  & 75\% Dropped \\ \midrule
ResNet50 & 0.32$\pm$0.16  & 0.13$\pm$0.11 & 0.07$\pm$0.09           \\
TnT-S    & 0.83$\pm$0.08  & 0.67$\pm$0.12 & 0.46$\pm$0.17           \\
ViT-L    & \textbf{0.92$\pm$0.06}  & \textbf{0.81$\pm$0.13}          & 0.50$\pm$0.21   \\
Deit-B   & 0.90$\pm$0.06  & 0.77$\pm$0.10 & \textbf{0.56$\pm$0.15}           \\
T2T-24   & 0.80$\pm$0.10  & 0.60$\pm$0.15 & 0.31$\pm$0.17           \\
\bottomrule
\end{tabular}}
\vspace{0.5em}
\captionof{table}{\small Correlation coefficient b/w features/final class tokens of original and occluded images for Random PatchDrop. Averaged across the ImageNet val. set.}
\label{tbl:correlation_coefficint_val_set}
\end{minipage}
\hspace{0.02\linewidth}
\begin{minipage}{0.48\linewidth}
\vspace{-2em}
\begin{subfigure}[H]{\linewidth}
\centering
\includegraphics[width=0.95\textwidth]{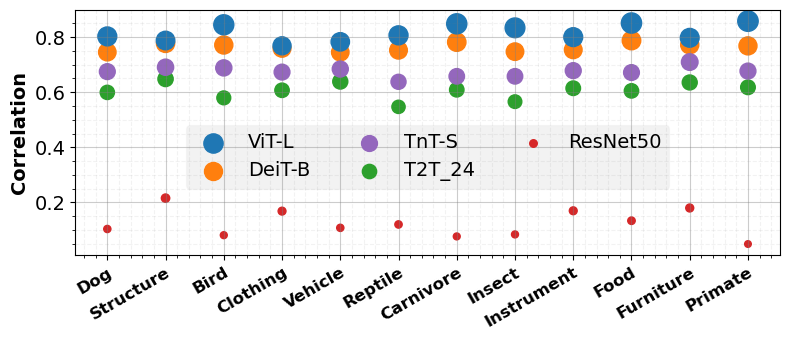}
\vspace{-0.5em}
\end{subfigure}
\captionof{figure}{\small Correlation b/w features/final tokens of original and occluded images for 50\% Random Drop. Results are averaged across classes for each superclass.}
\label{fig:correlation_coefficint_classwise_val_set}
\end{minipage}
\end{table}


\subsection{Shape vs. Texture: Can Transformer Model Both Characteristics?}
\label{sec:shape_bias}
Geirhos \etal \cite{geirhos2018imagenet} study shape vs. texture hypothesis and propose a training framework to enhance shape-bias in CNNs. 
We first carry out similar analysis and show that ViT models preform with a shape-bias much stronger than that of a CNN, and comparably to the ability of human visual system in recognizing shapes. 
However, this approach results in a significant drop in accuracy on the natural images. 
To address this issue, we introduce a shape token into the transformer architecture that learns to focus on shapes, thereby modeling both shape and texture related features within the same architecture using a distinct set of tokens. 
As such, we distill the shape information from a pretrained CNN model with high shape-bias \cite{geirhos2018imagenet}. 
Our distillation approach makes a balanced trade-off between high classification accuracy and strong shape-bias compared to the original ViT model. 

We outline both approaches below. 
Note that the measure introduced in \cite{geirhos2018imagenet} is used to quantify shape-bias within ViT models and compare against their CNN counterparts.

\textbf{Training without Local Texture:}
In this approach, we first remove local texture cues from the training data by creating a stylized version of ImageNet \cite{geirhos2018imagenet} named SIN. 
We then train tiny and small DeiT models \cite{touvron2020deit} on this dataset. 
Typically, ViTs use heavy data augmentations during training \cite{touvron2020deit}. 
However, learning with SIN is a difficult task due to less texture details and applying further augmentations on stylized samples distorts shape information and makes the training unstable. 
Thus, we train models on SIN without applying any augmentation, label smoothing or mixup.  

\begin{figure}[t]
\centering
\begin{subfigure}[b]{0.47\textwidth}
    \centering
    \includegraphics[width=\linewidth]{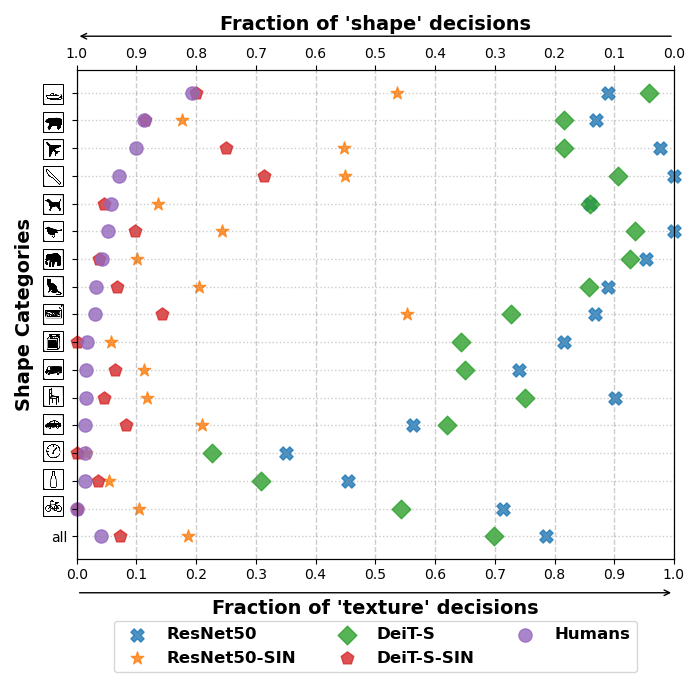}
    \vspace{-0.5cm}
\end{subfigure}
\hfill
\begin{minipage}[b]{0.51\textwidth}
  \begin{subfigure}[b]{0.95\linewidth}
    \centering
    \includegraphics[width=\linewidth]{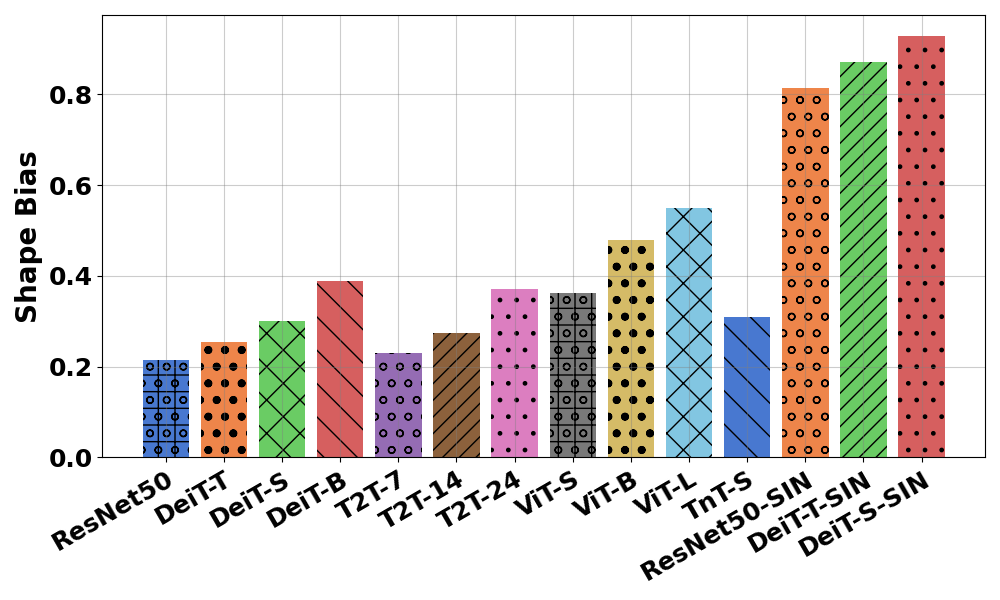}
    \label{fig:shape_bias_all}
  \end{subfigure}
\vspace{-0.8cm}
\caption{\emph{Shape-bias Analysis:} Shape-bias is defined as the fraction of correct decisions based on object shape. \emph{(Left)} Plot shows shape-texture tradeoff for CNN, ViT and Humans across different object classes. \emph{(Right)} class-mean shape-bias comparison. Overall, ViTs perform better than CNN. The shape bias increases significantly when trained on stylized ImageNet (SIN).}
\label{fig:shape_bias_analysis}
\end{minipage}
\end{figure}

We note that ViTs trained on ImageNet exhibit higher shape-bias in comparison to similar capacity CNN models \emph{e.g.,} DeiT-S (22-Million params) performs better than ResNet50 (23-Million params) (Fig.~\ref{fig:shape_bias_analysis}, \emph{right} plot). 
In contrast, the SIN trained ViTs consistently perform better than CNNs.
Interestingly, DeiT-S \cite{touvron2020deit} reaches human-level performance when trained on a SIN (Fig.~\ref{fig:shape_bias_analysis}, \emph{left} plot). 

\begin{SCtable}[][t]\setlength{\tabcolsep}{5pt}
\centering
\scalebox{0.75}{
\begin{tabular}{l|c|c|c|c}
\toprule
Model & Distilled& Token Type & ImageNet top-1 (\%) &Shape Bias\\
\midrule
DeiT-T-SIN &\xmark &cls & 40.5 & 0.87 \\
DeiT-T-SIN & \cmark&cls & 71.8 & 0.35\\
DeiT-T-SIN & \cmark&shape & 63.4 & 0.44\\
\midrule
DeiT-S-SIN &\xmark &cls & 52.5 & 0.93\\
DeiT-S-SIN & \cmark&cls & 75.3 & 0.39\\
DeiT-S-SIN & \cmark&shape & 67.7 & 0.47\\
\bottomrule
\end{tabular}
}
\vspace{0.5em}
\caption{Performance comparison of models trained on SIN. ViT produces dynamic features that can be controlled by auxiliary tokens. `cls' represents the class token. During distillation cls and shape tokens converged to vastly different solution using the same features as compared to \cite{touvron2020deit}.}
\label{tbl:shape_bias_distill}
\end{SCtable}

\begin{wrapfigure}[12]{r}{0.4\linewidth}
\centering
\vspace{-0.8em}
\includegraphics[width=1\linewidth]{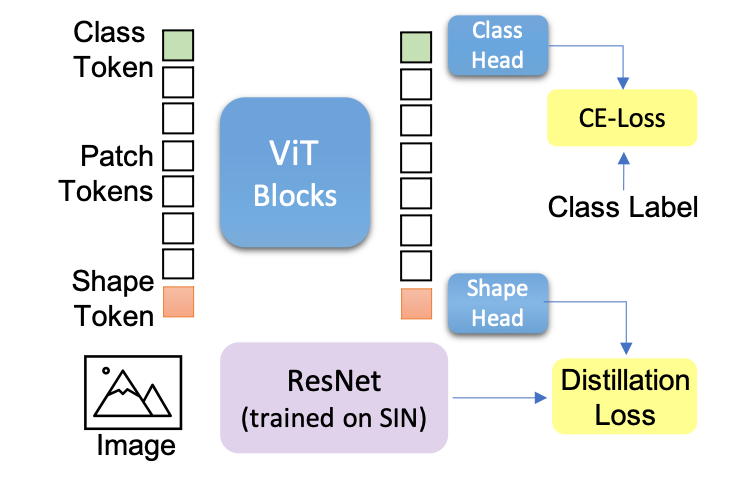}
\caption{Shape Distillation.}
\label{fig:shape_distill_block}
\vspace{0.8em}
\end{wrapfigure}

\textbf{Shape Distillation:}  
Knowledge distillation allows to compress large teacher models into smaller student models \cite{hinton2015distilling} as the teacher provides guidance to the student through soft labels.
We introduce a new shape token and adapt attentive distillation \cite{touvron2020deit} to distill shape knowledge from a CNN trained on the SIN dataset (ResNet50-SIN \cite{geirhos2018imagenet}). 
We observe that ViT features are dynamic in nature and can be controlled by auxiliary tokens to focus on the desired characteristics. 
This means that a single ViT model can exhibit both high shape and texture bias at the same time with separate tokens (Table~\ref{tbl:shape_bias_distill}). 
We achieve more balanced performance for classification as well as shape-bias measure when the shape token is introduced (Fig.~\ref{fig:shape_distill_block}). 
To demonstrate that these distinct tokens (for classification and shape) indeed model unique features, we compute cosine similarity (averaged over ImageNet val. set) between class and shape tokens of our distilled models, DeiT-T-SIN and DeiT-S-SIN, which turns out to be 0.35 and 0.68, respectively. 
This is significantly lower than the similarity between class and distillation tokens \cite{touvron2020deit}; 0.96 and 0.94 for DeiT-T and Deit-S, respectively. 
This confirms our hypothesis on modeling distinct features with separate tokens within ViTs, a unique capability that cannot be straightforwardly achieved with CNNs.
Further, it offers other benefits as we explain next.

\textbf{Shape-biased ViT Offers Automated Object Segmentation:}  Interestingly, training without local texture or with shape distillation allows a ViT to concentrate on foreground objects in the scene and ignore the background (Table~\ref{tab:shape_bias_attention}, Fig.~\ref{fig:shape_bias_attention}). 
This offers an automated semantic segmentation for an image although the model is never shown pixel-wise object labels. 
That is, shape-bias can be used as self-supervision signals for the ViT model to learn distinct shape-related features that help localize the right foreground object. 
We note that a ViT trained without emphasis on shape does not perform well (Table~\ref{tab:shape_bias_attention}).

\begin{table}[t]\setlength{\tabcolsep}{3.5pt}
\begin{minipage}{0.49\linewidth}
\centering
\scalebox{0.85}{
\begin{tabular}{l|c|c|c}
\toprule
Model & Distilled& Token Type & Jaccard Index\\
\midrule
DeiT-T-Random & \xmark  & cls        & 19.6\\
DeiT-T &        \xmark  & cls        & 32.2\\
DeiT-T-SIN &    \xmark  & cls        & 29.4\\
DeiT-T-SIN &    \cmark  & cls        & 40.0\\
DeiT-T-SIN &    \cmark  & shape      & 42.2\\
\midrule
DeiT-S-Random & \xmark  & cls        & 22.0\\
DeiT-S &        \xmark  & cls        & 29.2\\
DeiT-S-SIN &    \xmark  & cls        & 37.5\\
DeiT-S-SIN &    \cmark  & cls        & 42.0\\
DeiT-S-SIN &    \cmark  & shape      & 42.4\\
\bottomrule
\end{tabular}
}
\vspace{0.5em}
\caption{We compute the Jaccard similarity between ground truth and masks generated from the attention maps of ViT models (similar to \cite{caron2021emerging} with threshold 0.9) over the PASCAL-VOC12 validation set. Only class level ImageNet labels are used for training these models. Our results indicate that supervised ViTs can be used for automated segmentation and perform closer to the self-supervised method DINO \cite{caron2021emerging}.}
\label{tab:shape_bias_attention}
\end{minipage}
\hspace{0.04\linewidth}
\begin{minipage}{0.49\linewidth}
\begin{subfigure}[b]{\linewidth}
\begin{subfigure}[b]{0.32\linewidth}        
    \centering
    \includegraphics[width=\linewidth, height=1.9cm]{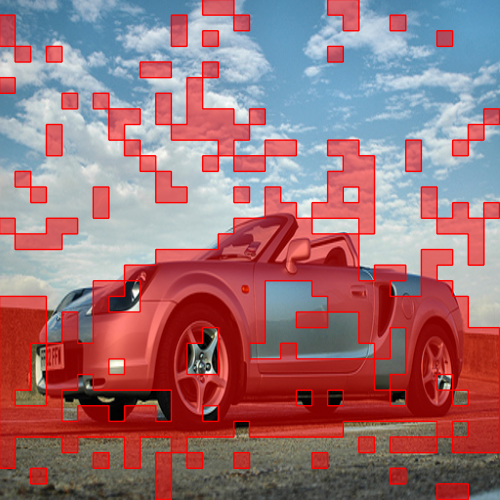}
\end{subfigure}
\begin{subfigure}[b]{0.32\linewidth}        
    \centering
    \includegraphics[width=\linewidth, height=1.9cm]{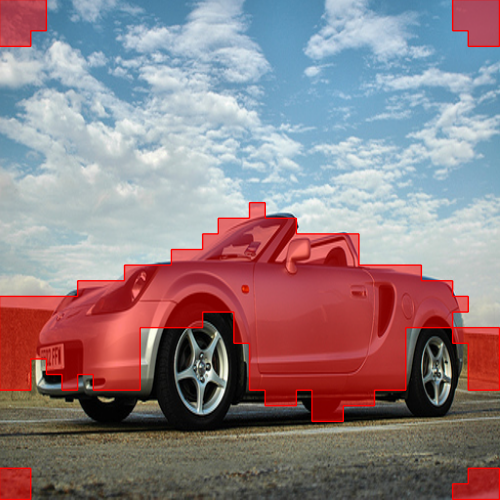}
\end{subfigure}
\begin{subfigure}[b]{0.32\linewidth}        
    \centering
    \includegraphics[width=\linewidth, height=1.9cm]{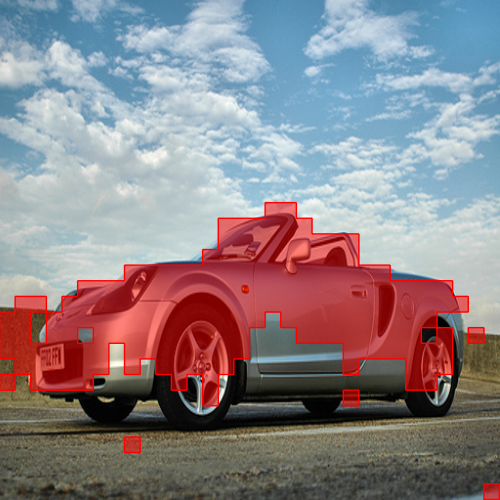}
\end{subfigure}
\end{subfigure}
\begin{subfigure}[b]{\linewidth}
\begin{subfigure}[b]{0.32\linewidth}        
    \centering
    \includegraphics[width=\linewidth, height=1.9cm]{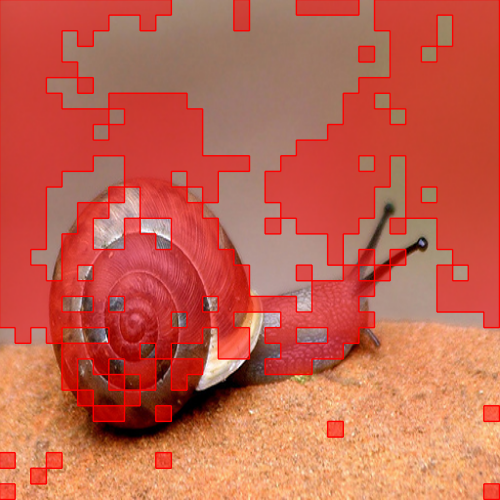}
\end{subfigure}
\begin{subfigure}[b]{0.32\linewidth}        
    \centering
    \includegraphics[width=\linewidth, height=1.9cm]{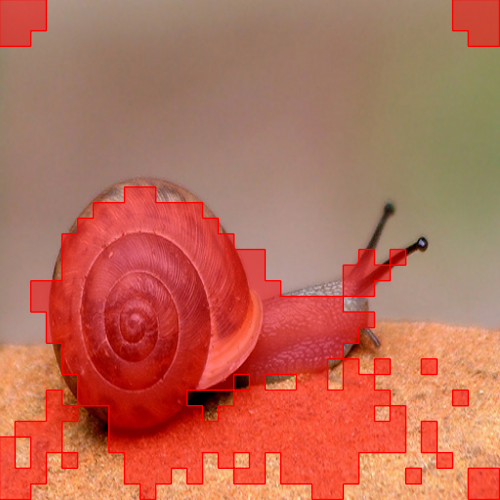}
\end{subfigure}
\begin{subfigure}[b]{0.32\linewidth}        
    \centering
    \includegraphics[width=\linewidth, height=1.9cm]{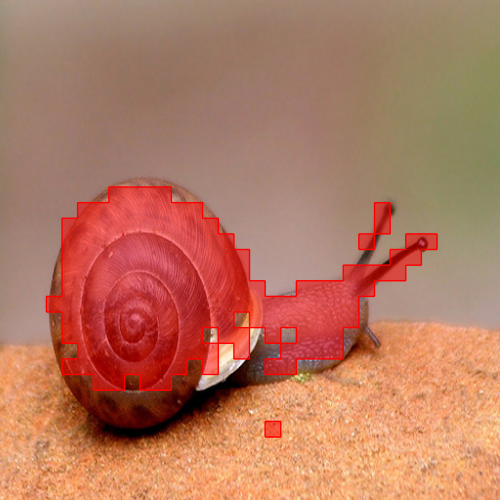}
\end{subfigure}
\end{subfigure}
\begin{subfigure}[b]{\linewidth}
\begin{subfigure}[b]{0.32\linewidth}        
    \centering
    \includegraphics[width=\linewidth, height=1.9cm]{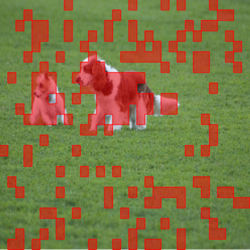} 
    \caption*{\tiny\textbf{DeiT-S}}
\end{subfigure}
\begin{subfigure}[b]{0.32\linewidth}        
    \centering
    \includegraphics[width=\linewidth, height=1.9cm]{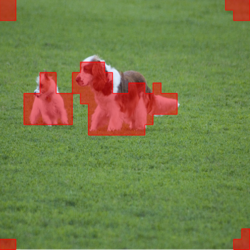} 
    \caption*{\tiny\textbf{DeiT-S-SIN}}
\end{subfigure}
\begin{subfigure}[b]{0.32\linewidth}        
    \centering
    \includegraphics[width=\linewidth, height=1.9cm]{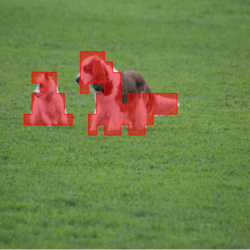} 
    \caption*{\tiny\textbf{DeiT-S-SIN (Distilled)}}
\end{subfigure}
\vspace{-0.5em}
\end{subfigure}
\captionof{figure}{Segmentation maps from ViTs. Shape distillation performs better than standard supervised models.} 
\label{fig:shape_bias_attention}
\end{minipage}
\vspace{-1.5em}
\end{table}

The above results show that properly trained ViT models offer shape-bias nearly as high as the human's ability to recognize shapes. 
This leads us to question if positional encoding is the key that helps ViTs achieve high performance under severe occlusions (as it can potentially allow later layers to recover the missing information with just a few image patches given their spatial ordering). 
This possibility is examined next. 

\subsection{Does Positional Encoding Preserve the Global Image Context?}
\label{sec:shuffle}
\begin{figure}[b!]
\begin{minipage}{.65\textwidth}
    \begin{subfigure}[b]{0.48\linewidth}        
        \centering
        \includegraphics[width=\linewidth]{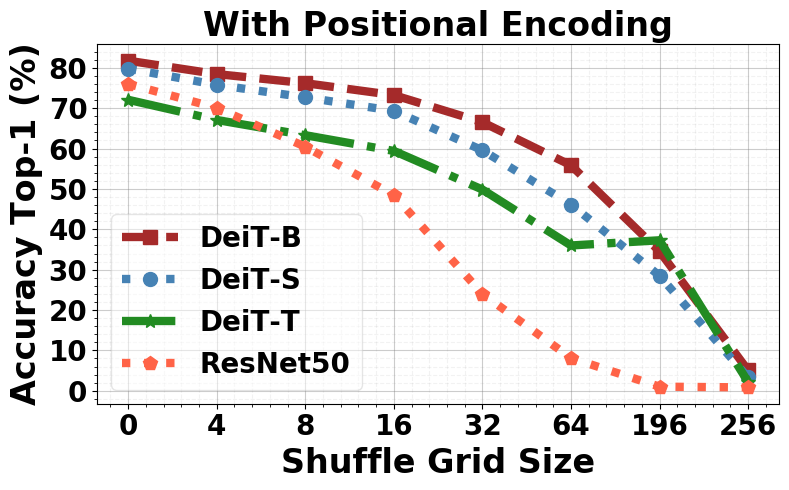}
    \end{subfigure}
    \begin{subfigure}[b]{0.48\linewidth}        
        \centering
        \includegraphics[width=\linewidth]{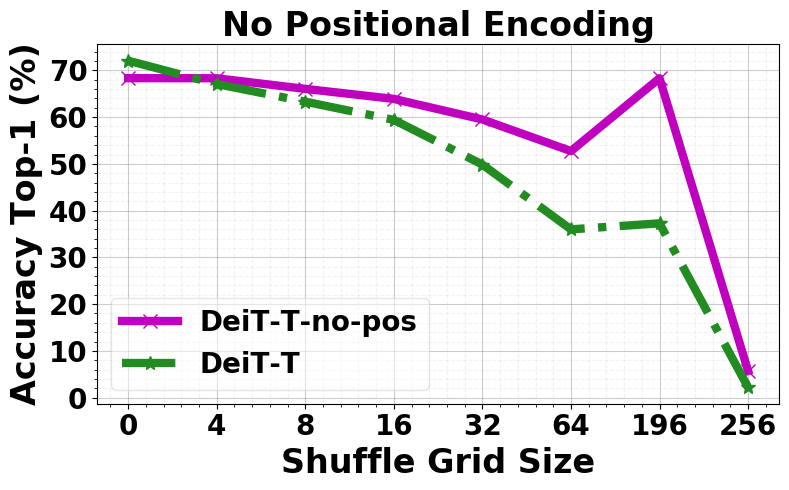}
    \end{subfigure}
    \caption{\small Models trained on 196 image patches.  Top-1 (\%) accuracy over ImageNet val. set when patches are shuffled. Note the performance peaks when shuffle grid size is equal to the original number of patches used during training, since it equals to only changing the position of input patch (and not disturbing the patch content).}
    \label{fig:shuffle_with_without_positional_encoding}
\end{minipage}
\hfill
\begin{minipage}{.3\textwidth}
    \begin{subfigure}[b]{\linewidth}        
        \centering
        \includegraphics[width=\linewidth]{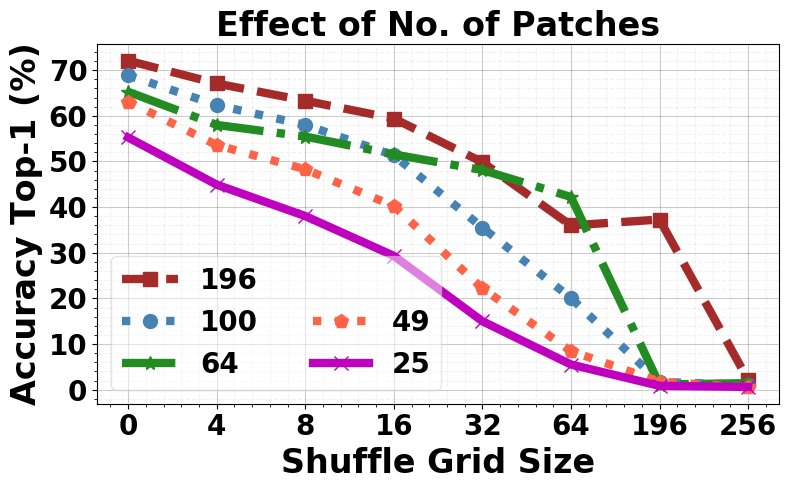}
    \end{subfigure}
    \vspace{-1.5em}
\caption{DeiT-T \cite{touvron2020deit} trained on different number of image patches. Reducing patch size decreases the overall performance but also increases sensitivity to shuffle grid size.}
\label{fig:shuffle_with_different_no_of_patches}
\end{minipage}    
\end{figure}

Transformers' ability to process long-range sequences in parallel using self-attention \cite{vaswani2017attention} (instead of a sequential design in RNN~\cite{hochreiter1997long}) is invariant to sequence ordering. %
For images, the order of patches represents the overall image structure and global composition. 
Since ViTs operate on a sequence of images patches, changing the order of sequence \emph{e.g.,} shuffling the patches can destroy the image structure. 
Current ViTs \cite{dosovitskiy2020image, touvron2020deit, yuan2021tokens, mao2021transformer} use positional encoding to preserve this context. 
Here, we analyze if the sequence order modeled by positional encoding allows ViT to excel under occlusion handling.
Our analysis suggests that transformers show high permutation invariance to the patch positions, and the effect of positional encoding towards injecting structural information of images to ViT models is limited (Fig.~\ref{fig:shuffle_with_without_positional_encoding}). 
This observation is consistent with the findings in the language domain \cite{irie2019language} as described below.

\begin{wrapfigure}[8]{r}{0.6\textwidth}
    \centering
        \centering
        \vspace{-1.5em}
        \includegraphics[width=\linewidth]{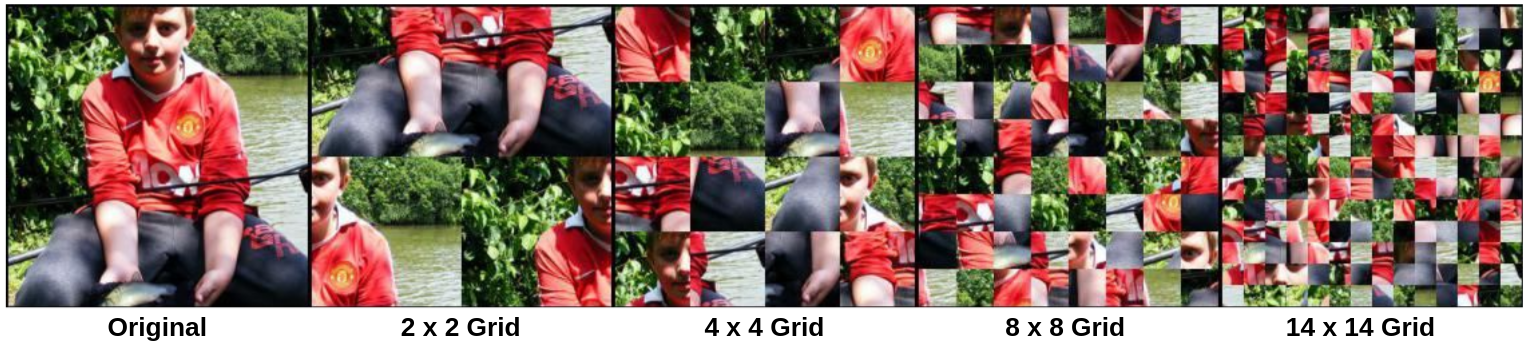}
        \vspace{-1.5em}
         \caption{\small An illustration of shuffle operation applied on images used to eliminate their structural information. (\emph{best viewed zoomed-in})}
    \label{fig:shuffle_demo}\vspace{-0em}
\end{wrapfigure}

\textbf{Sensitivity to Spatial Structure:}
We remove the structural information within images (spatial relationships) as illustrated in Fig.~\ref{fig:shuffle_demo} by defining a shuffling operation on input image patches.
Fig.~\ref{fig:shuffle_with_without_positional_encoding} shows that the DeiT models \cite{touvron2020deit} retain accuracy better than their CNN counterparts when spatial structure of input images is disturbed.
This also indicates that positional encoding is not absolutely crucial for right classification decisions, and the model does not ``recover'' global image context using the patch sequence information preserved in the positional encodings. 
Without encoding, the ViT performs reasonably well and achieves better permutation invariance than a ViT using position encoding
 (Fig.~\ref{fig:shuffle_with_without_positional_encoding}). 
Finally, when the patch size is varied during ViT training, the permutation invariance property is also degraded along with the accuracy on unshuffled natural images (Fig.~\ref{fig:shuffle_with_different_no_of_patches}). 
Overall, we attribute the permutation invariance performance of ViTs to their dynamic receptive field that depends on the input patch and can adjust attention with the other sequence elements such that moderately shuffling the elements does not degrade the performance significantly.

The above analysis shows that just like the texture-bias hypothesis does not apply to ViTs, the dependence on positional encodings to perform well under occlusions is also incorrect. 
This leads us to the conclude that ViTs robustness is due to its flexible and dynamic receptive field (see Fig.~\ref{fig:patchdrop_attention_maps}) which depends on the content of an input image. 
We now delve further deep into the robustness of ViT, and study its performance under adversarial perturbations and common corruptions. 

\subsection{Robustness of Vision Transformers to Adversarial and Natural Perturbations}
After analyzing the ability of ViTs to encode shape information (Sec.~\ref{sec:shape_bias}), one ensuing question is:
\emph{Does higher shape-bias help achieve better robustness?} 
In Table~\ref{tab:common_corruptions}, we investigate this by calculating mean corruption error (mCE) \cite{hendrycks2019benchmarking} on a variety of synthetic common corruptions (\emph{e.g.,} rain, fog, snow and noise). 
A ViT with similar parameters as CNN (\emph{e.g.,} DeiT-S) is more robust to image corruptions than ResNet50 trained with augmentations (Augmix \cite{hendrycks2020augmix}). 
Interestingly, CNNs and ViTs trained without augmentations on ImageNet or SIN are more vulnerable to corruptions. 
These findings are consistent with \cite{mummadi2021does}, and suggest that augmentations improve robustness against common corruptions.

\setcounter{table}{3}
\begin{table}[t!]\setlength{\tabcolsep}{3.5pt}
\centering \small
\scalebox{0.9}{
\begin{tabular}{cccccc|cccc}
\toprule
\multicolumn{6}{c}{Trained with Augmentations} & \multicolumn{4}{c}{Trained without Augmentation} \\ \toprule
\texttt{DeiT-B} & \texttt{DeiT-S} & \texttt{DeiT-T} & \texttt{T2T-24} & \texttt{TnT-S} & \texttt{Augmix} & \texttt{ResNet50} & \texttt{ResNet50-SIN} & \texttt{DeiT-T-SIN} & \texttt{DeiT-S-SIN}\\
\midrule
 48.5 & 54.6 & 71.1 & 49.1 & 53.1 & 65.3 & 76.7 & 77.3 & 94.4 & 84.0\\
\bottomrule
\end{tabular}
}\vspace{0.3em}
\caption{mean Corruption Error (mCE) across common corruptions \cite{hendrycks2019benchmarking} (lower the better). While ViTs have better robustness compared to CNNs, training to achieve a higher shape-bias makes both CNNs and ViTs more vulnerable to natural distribution shifts. All models trained with augmentations (ViT or CNN) have lower mCE in comparison to models trained without augmentations on ImageNet or SIN.}
\label{tab:common_corruptions}
\end{table}
\begin{figure}[t!]
\centering
\begin{minipage}{.66\textwidth}
  \begin{minipage}{.48\textwidth}
  	\centering
    \includegraphics[ width=\linewidth]{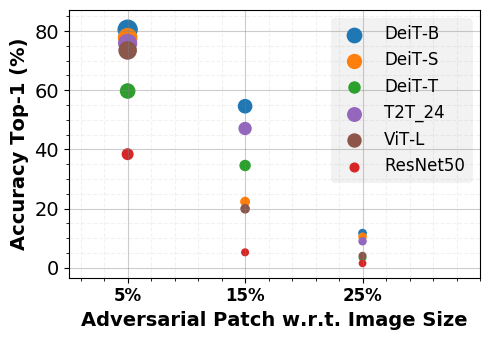}\
  \end{minipage}
    \begin{minipage}{.48\textwidth}
	\centering
    \includegraphics[width=\linewidth]{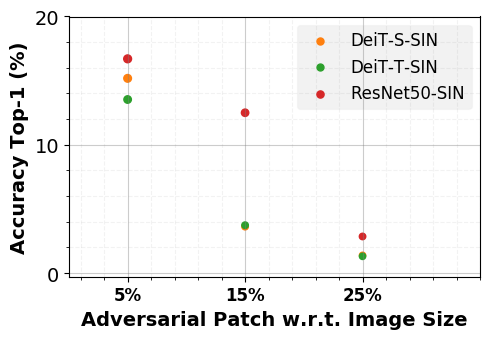}\
  \end{minipage}
\end{minipage}
 \hfill
\begin{minipage}{.33\textwidth}
\captionof{figure}{\muz{Robustness against adversarial patch attack. ViTs even with less parameters exhibit a higher robustness than CNN. Models trained on ImageNet are more robust than the ones trained on SIN. Results are averaged across five runs of patch attack over ImageNet val. set.}}
\label{fig:adv_plot}
\end{minipage}
\end{figure}

\begin{figure}[t!]
\centering
  \begin{minipage}{.24\textwidth}
  	\centering
    \includegraphics[ width=\linewidth]{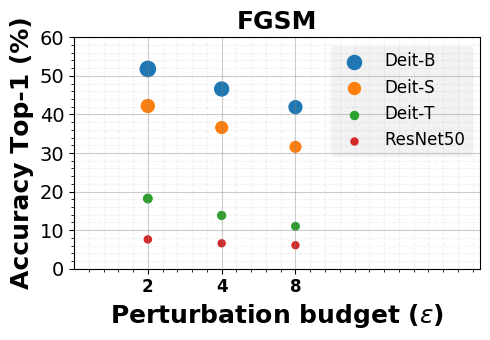}\
  \end{minipage}
    \begin{minipage}{.24\textwidth} 
	\centering
    \includegraphics[width=\linewidth]{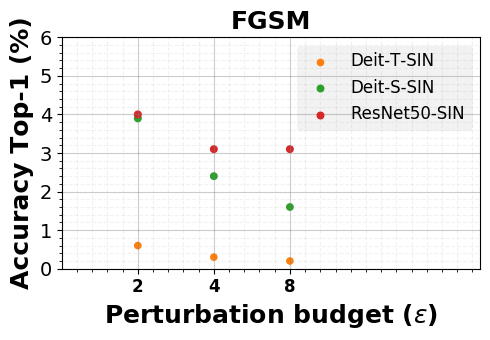}\
  \end{minipage}
    \begin{minipage}{.24\textwidth}
  	\centering
    \includegraphics[ width=\linewidth]{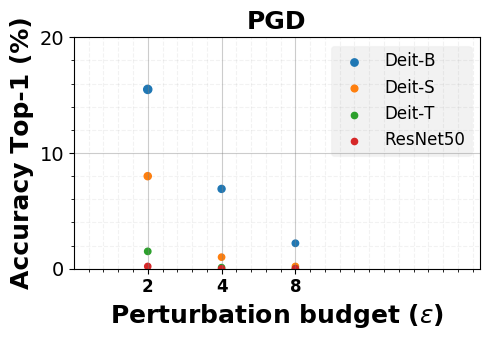}\
  \end{minipage}
    \begin{minipage}{.24\textwidth} 
	\centering
    \includegraphics[width=\linewidth]{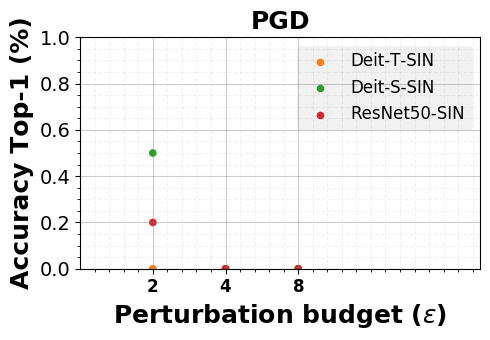}\
  \end{minipage}
\caption{\muz{Robustness against sample specific attacks including single step, FGSM \cite{goodfellow2014explaining}, and multi-step, PGD \cite{madry2017towards}. ViTs even with less parameters exhibit a higher robustness than CNN. PGD ran for 5 iterations only. Attacks are evaluated under $l_\infty$ norm and $\epsilon$ represents the perturbation budget by which each pixel is changed in the input image. Results are reported over the ImageNet val. set.}}
\label{fig:adv_plot_iterative_attacks}

\end{figure}



%
%
\muz{We observe similar performance against untargeted, universal adversarial patch attack \cite{brown2017adversarial} and sample specific attacks including single step, fast gradient sign method (FGSM) \cite{goodfellow2014explaining}, and multi-step projected gradient attack known as PGD \cite{madry2017towards}.  Adversarial patch attack \cite{brown2017adversarial} is unbounded that is it can change pixel values at certain location in the input image by any amount, while sample specific attacks \cite{goodfellow2014explaining, madry2017towards} are bounded by $l_\infty$ norm with a perturbation budget $\epsilon$, where $\epsilon$  represents the amount by which each pixel is changed in the entire image.}
ViTs and CNN trained on SIN are significantly more vulnerable to adversarial attack than models trained on ImageNet (Figs.~\ref{fig:adv_plot} and ~\ref{fig:adv_plot_iterative_attacks}), due to the shape-bias vs. robustness trade-off \cite{mummadi2021does}.

Given the strong robustness properties of ViT as well as their representation capability in terms of shape-bias, automated segmentation and flexible receptive field, we analyze their utility as an off-the-shelf feature extractor to replace CNNs as the default feature extraction mechanism \cite{Razavian2014}. 

\vspace{1em}
\subsection{Effective Off-the-shelf Tokens for Vision Transformer}
\label{sec:off_shelf}
\begin{figure}[t!]
\begin{minipage}{.53\textwidth}
    \begin{subfigure}[b]{\linewidth}        
        \centering
        \includegraphics[width=0.98\linewidth]{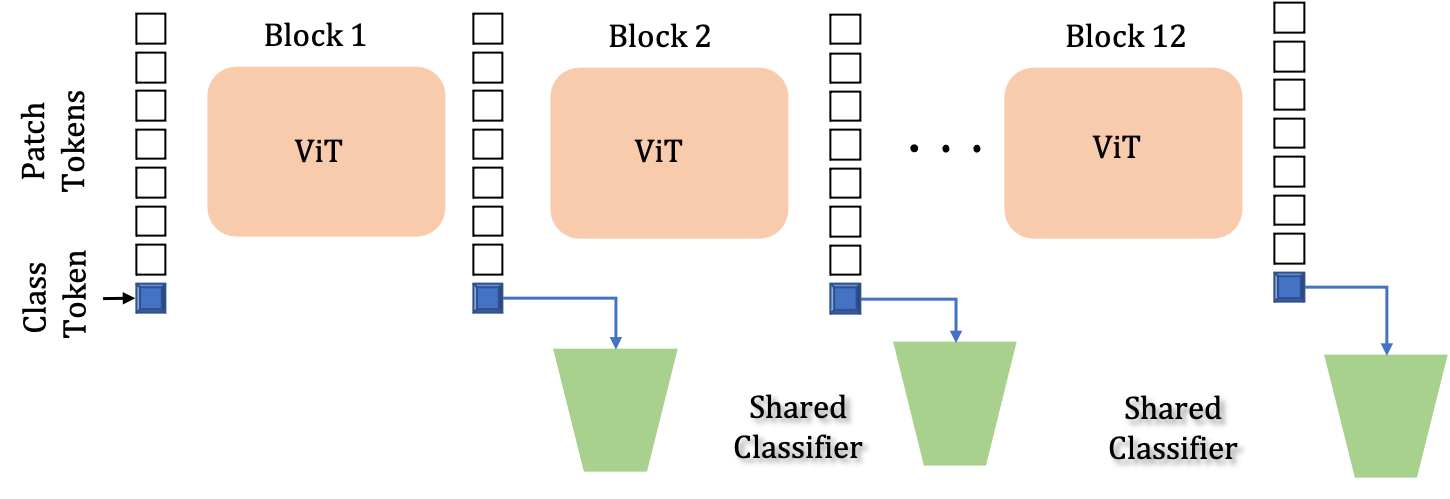}
    \end{subfigure}
    \caption{A single ViT model can provide a features ensemble since class token from each block can be processed by the classifier independently. This allows us to identify the most discriminative tokens useful for transfer learning.}
    \label{fig:enemble_method}
\end{minipage}
\hfill
\begin{minipage}{.45\textwidth}
    \begin{subfigure}[b]{\linewidth}        
        \centering
        \includegraphics[width=\linewidth]{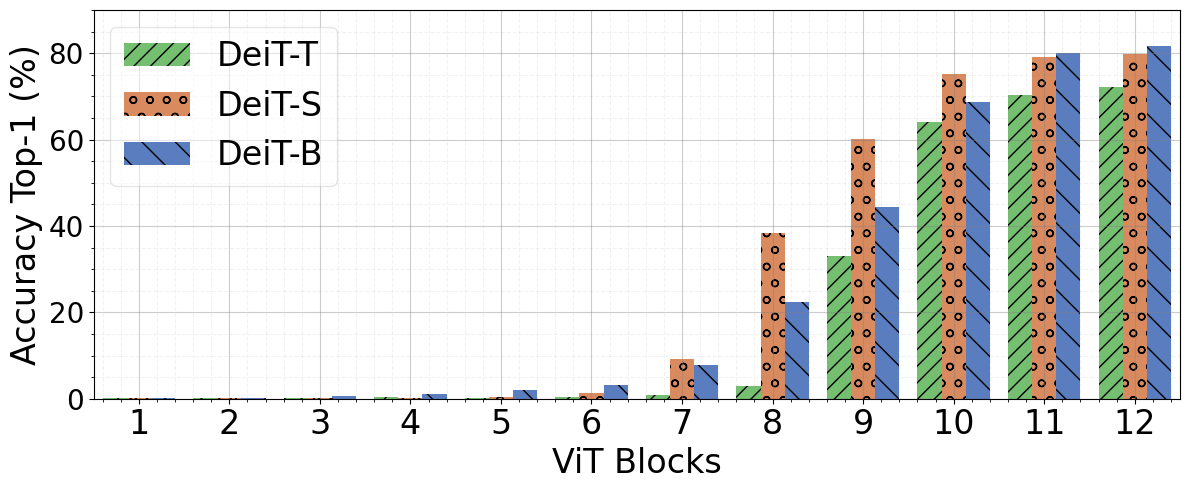}
    \end{subfigure}
\caption{ Top-1 (\%) for ImageNet val. set for class tokens produced by each ViT block.  Class tokens from the last few layers exhibit highest performance indicating the most discriminative tokens.}
\label{fig:blockwise_acc}
\end{minipage}    
\end{figure}

A unique characteristic of ViT models is that each block within the model generates a class token which can be processed by the classification head separately (Fig. \ref{fig:enemble_method}). 
This allows us to measure the discriminative ability of each individual block of an ImageNet pre-trained ViT as shown in Fig. \ref{fig:blockwise_acc}. 
Class tokens generated by the deeper blocks are more discriminative and we use this insight to identify an effective ensemble of blocks whose tokens have the best downstream transferability. 

\begin{SCtable}[][t]\setlength{\tabcolsep}{3pt}
\centering
\scalebox{0.85}{
\small
\begin{tabular}{l c c c c c}
\toprule
Blocks & Class  & Patch  & CUB  & Flowers  & iNaturalist  \\
       & Tokens       & Tokens        &   \cite{cub}  &      \cite{Nilsback08}   & \cite{van2018inaturalist}\\
\midrule
\multirow{2}{*}{Only 12$^{th}$ (last block)}& \cmark & \xmark & 68.16 & 82.58 & 38.28\\ 
& \cmark & \cmark &  70.66 & 86.58 &	41.22 \\
\midrule
\multirow{2}{*}{From 1$^{st}$ to 12$^{th}$}& \cmark & \xmark &  72.90& \textbf{91.38} & 44.03 \\ 
& \cmark & \cmark & 73.16&  91.27&	43.33\\
\midrule
\multirow{2}{*}{From 9$^{th}$ to 12$^{th}$}& \cmark & \xmark & \textbf{73.58} &  90.00 & \textbf{45.15}\\ 
& \cmark & \cmark & 73.37&90.33 & 45.12 \\
\bottomrule
\end{tabular}
}
\vspace{0.5em}
\caption{\muz{Ablative Study for off-the-shelf feature transfer on three datasets using ImageNet pretrained DeiT-S \cite{touvron2020deit}. A linear classifier is learned on only a concatenation of class tokens or the combination of class and averaged patch tokens at various blocks. We note class token from blocks 9-12 are most discriminative (Fig.~\ref{fig:blockwise_acc}) and have the highest  transferability in terms of Top-1 (\%) accuracy.}}
\label{tbl:off_shelf_ablative}
\end{SCtable}

\textbf{Transfer Methodology:} 
As illustrated in Fig.~\ref{fig:blockwise_acc}, we analyze the block-wise classification accuracy of DeiT models and determine the discriminative information is captured within the class tokens of the last few blocks. 
\muz{As such, we conduct an ablation study for off-the-shelf transfer learning on fine-grained classification dataset CUB \cite{cub}, Flowers \cite{Nilsback08}  and large scale iNaturalist \cite{van2018inaturalist} using  DeiT-S \cite{touvron2020deit} as reported in Table~\ref{tbl:off_shelf_ablative}.} 
Here, we concatenate the class tokens (optionally combined with average patch tokens) from different blocks and train a linear classifier to transfer the features to downstream tasks. 
Note that a patch token is generated by averaging along the patch dimension. 
The scheme that concatenate class tokens from the last four blocks shows the best transfer learning performance. 
We refer to this transfer methodology as DeiT-S (ensemble).
\muz{Concatenation of both class and averaged patch tokens from all blocks helps achieve similar performance compared to the tokens from the last four blocks but requires significantly large parameters to train. We find some exception to this on the Flower dataset \cite{Nilsback08} where using class tokens from all blocks have relatively better improvement (only 1.2\%), compared to the class tokens from the last four blocks (Table ~\ref{tbl:off_shelf_ablative}). However, concatenating tokens from all blocks also increases the number of parameters e.g., transfer to Flowers from all tokens has 3 times more learnable parameters than using only the last four tokens.}
We conduct further experimentation with DeiT-S (ensemble) across a broader range of tasks to validate our hypothesis. 
We further compare against a pre-trained ResNet50 baseline, by using features before the logit layer.

\begin{figure}[t!]
    \centering
    \begin{subfigure}[b]{0.48\linewidth}        
        \centering
        \includegraphics[width=\linewidth]{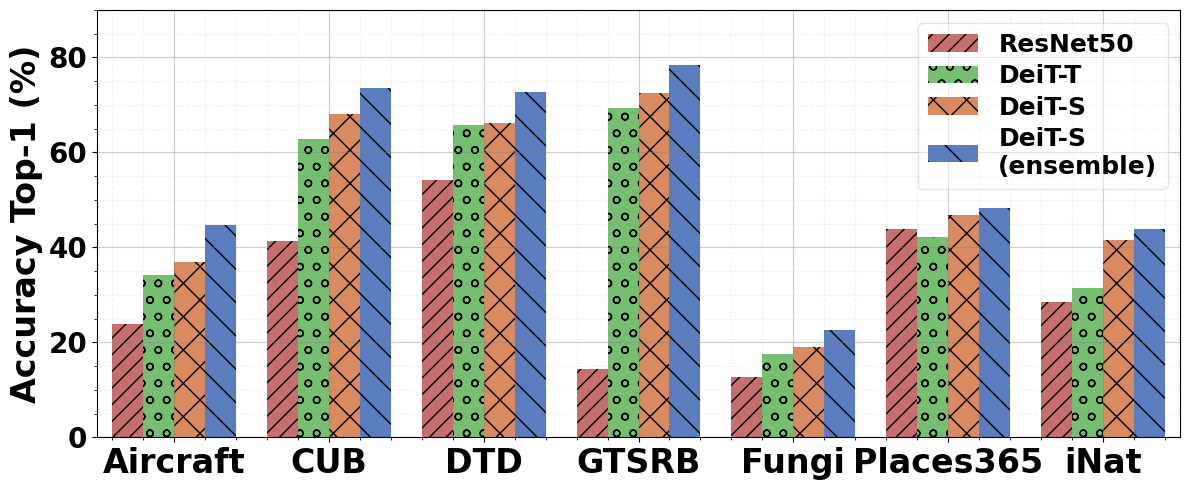}
    \end{subfigure}
    \begin{subfigure}[b]{0.48\linewidth}        
        \centering
        \includegraphics[width=\linewidth]{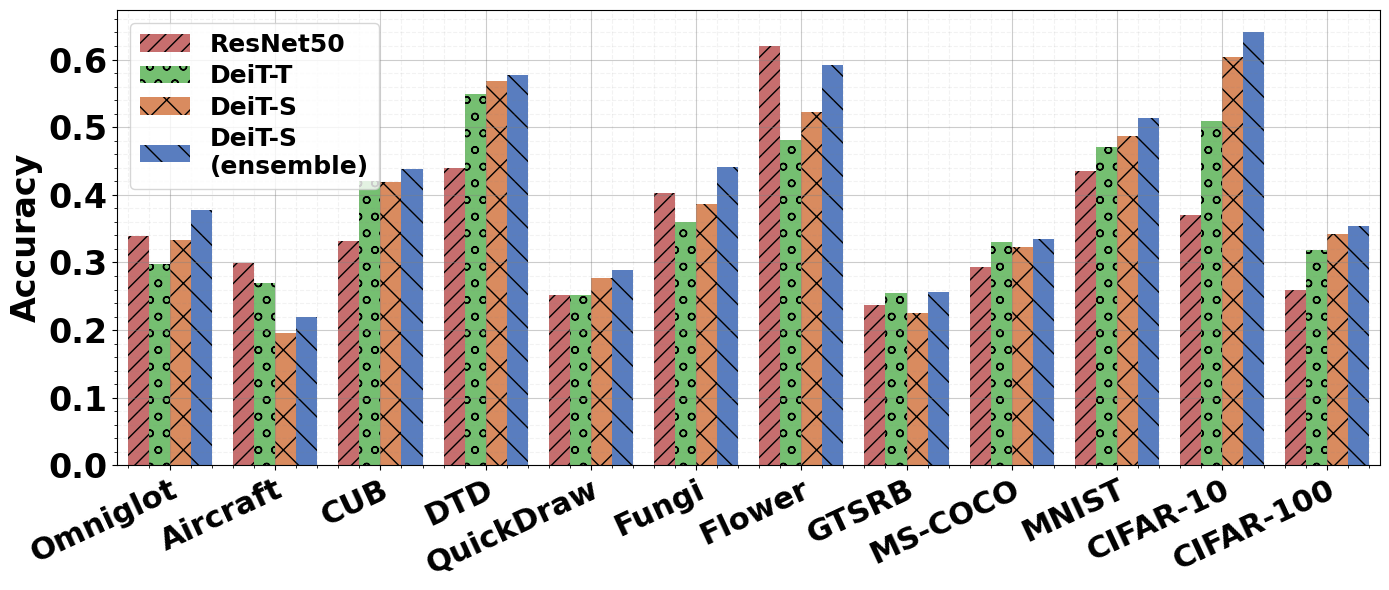}
    \end{subfigure}
     \caption{\muz{Off-the-shelf ViT features transfer better than CNNs. We explore transferability of learned representations using generic classification as well as few-shot classification for out-of-domain tasks. In the case of classification (\emph{left}), the ImageNet pre-trained ViTs transfer better than their CNN counterparts across tasks. In the case of few-shot learning (\emph{right}), ImageNet pre-trained ViTs perform better on average.}}
     \label{fig:off_shelf}
\end{figure}

\texttt{Visual Classification:}
We analyze the transferability of off-the-shelf  features across several datasets including Aircraft \cite{aircraft}, CUB \cite{cub}, DTD \cite{dtd}, GTSRB \cite{gtsrb}, Fungi \cite{fungi_dataset}, Places365 \cite{zhou2017places} and iNaturalist~\cite{van2018inaturalist}.
These datasets are developed for fine-grained recognition, texture classification, traffic sign recognition, species classification and scene recognition with 100, 200, 47, 43, 1394, 365 and 1010 classes respectively. 
We train a linear classifier on top of the extracted features over the train split of each dataset, and evaluate the performance on their respective test splits.  
The ViT features show clear improvements over the CNN baseline (Fig.~\ref{fig:off_shelf}). 
We note that DeiT-T, which requires about 5 times fewer parameters than ResNet50, performs better among all datasets. 
Furthermore, the model with the proposed ensemble strategy achieves the best results across all datasets. 

\texttt{Few-Shot Learning:} We consider meta-dataset \cite{meta_dataset} designed as a
large-scale few-shot learning (FSL) benchmark containing a diverse set of datasets from multiple domains. This includes letters of alphabets, hand-drawn sketches, images of textures, and fine-grained classes making it a challenging dataset involving a domain adaption requirement as well. We follow the standard setting of training on ImageNet and testing on all other datasets which are considered as the downstream tasks.

\muz{In our experiments, we use a network pre-trained for classification on ImageNet dataset to extract features. For each downstream dataset, under the FSL setting, a support set of labelled images is available for every test query. We use the extracted features to learn a linear classifier over the support set for each query (similar to \cite{yonglong_rfs}), and evaluate using the standard FSL protocol defined in \cite{meta_dataset}. This evaluation involves a varying number of shots specific for each downstream dataset.  On average, the ViT features transfer better across these diverse domains (Fig.~\ref{fig:off_shelf}) in comparison to the CNN baseline. Furthermore, we note that the transfer performance of ViT is further boosted using the proposed ensemble strategy. We also highlight the improvement in QuickDraw, a dataset containing hand-drawn sketches, which aligns with our findings on improved shape-bias of ViT models in contrast to CNN models (see Sec.~\ref{sec:shape_bias} for elaborate discussion). }

\section{Discussion and Conclusions}\vspace{-0.8em}
In this paper, we analyze intriguing properties of ViTs in terms of robustness and generalizability. 
We test with a variety of ViT models on fifteen vision datasets. All the models are trained on 4 V100 GPUs.
We demonstrate favorable merits of ViTs over CNNs for occlusion handling, robustness to distributional shifts and patch permutations, automatic segmentation without pixel supervision, and robustness against adversarial patches, sample specific adversarial attacks and common corruptions. 
Moreover, we demonstrate strong transferability of off-the-shelf ViT features to a number of downstream tasks with the proposed feature ensemble from a single ViT model. 
An interesting future research direction is to explore how the diverse range of cues modeled within a single ViT using separate tokens can be effectively combined to complement each other. 
\muz{Similarly, we found that ViTs auto-segmentation property stems from their ability to encode shape information.  We believe that integrating our approach and DINO \cite{caron2021emerging} is worth exploring in the future. To highlight few open research questions: a) Can self-supervision on stylized ImageNet (SIN) improve segmentation ability of DINO?, and b) Can a modified DINO training scheme with texture (IN) based local views and shape (SIN) based global views enhance (and generalize) its auto-segmentation capability?} 

Our current set of experiments are based on ImageNet (ILSVRC'12) pre-trained ViTs, which pose the risk of reflecting potential biases in the learned representations. 
The data is mostly Western, and encodes several gender/ethnicity stereotypes with under-representation of certain groups \cite{yang2019fairer}.
This version of the ImageNet also poses privacy risks, due to the unblurred human faces. 
In future, we will use a recent ImageNet version which addresses the above issues \cite{yang2021study}. 

\clearpage
\section*{Acknowledgments}
This work is supported in part by NSF CAREER grant 1149783, and VR starting grant (2016-05543). M. Hayat is supported by Australian Research Council DECRA fellowship DE200101100.
{\small
\bibliographystyle{unsrt}
\bibliography{egbib}

\begin{thebibliography}{10}

\bibitem{khan2021transformers}
Salman Khan, Muzammal Naseer, Munawar Hayat, Syed~Waqas Zamir, Fahad~Shahbaz
  Khan, and Mubarak Shah.
\newblock Transformers in vision: A survey.
\newblock {\em arXiv preprint arXiv:2101.01169}, 2021.

\bibitem{dosovitskiy2020image}
Alexey Dosovitskiy, Lucas Beyer, Alexander Kolesnikov, Dirk Weissenborn,
  Xiaohua Zhai, Thomas Unterthiner, Mostafa Dehghani, Matthias Minderer, Georg
  Heigold, Sylvain Gelly, et~al.
\newblock An image is worth 16x16 words: Transformers for image recognition at
  scale.
\newblock {\em arXiv preprint arXiv:2010.11929}, 2020.

\bibitem{touvron2020deit}
Hugo Touvron, Matthieu Cord, Matthijs Douze, Francisco Massa, Alexandre
  Sablayrolles, and Herv\'e J\'egou.
\newblock Training data-efficient image transformers \& distillation through
  attention.
\newblock {\em arXiv preprint arXiv:2012.12877}, 2020.

\bibitem{yuan2021tokens}
Li~Yuan, Yunpeng Chen, Tao Wang, Weihao Yu, Yujun Shi, Francis~EH Tay, Jiashi
  Feng, and Shuicheng Yan.
\newblock Tokens-to-token vit: Training vision transformers from scratch on
  imagenet.
\newblock {\em arXiv preprint arXiv:2101.11986}, 2021.

\bibitem{ramachandran2019stand}
Prajit Ramachandran, Niki Parmar, Ashish Vaswani, Irwan Bello, Anselm Levskaya,
  and Jonathon Shlens.
\newblock Stand-alone self-attention in vision models.
\newblock {\em arXiv preprint arXiv:1906.05909}, 2019.

\bibitem{hu2019local}
Han Hu, Zheng Zhang, Zhenda Xie, and Stephen Lin.
\newblock Local relation networks for image recognition.
\newblock In {\em IEEE Conference on Computer Vision and Pattern Recognition},
  pages 3464--3473, 2019.

\bibitem{vaswani2021scaling}
Ashish Vaswani, Prajit Ramachandran, Aravind Srinivas, Niki Parmar, Blake
  Hechtman, and Jonathon Shlens.
\newblock Scaling local self-attention for parameter efficient visual
  backbones.
\newblock {\em arXiv preprint arXiv:2103.12731}, 2021.

\bibitem{ILSVRC15}
Olga Russakovsky, Jia Deng, Hao Su, Jonathan Krause, Sanjeev Satheesh, Sean Ma,
  Zhiheng Huang, Andrej Karpathy, Aditya Khosla, Michael Bernstein,
  Alexander~C. Berg, and Li~Fei-Fei.
\newblock {ImageNet Large Scale Visual Recognition Challenge}.
\newblock {\em International Journal of Computer Vision}, 115(3):211--252,
  2015.

\bibitem{geirhos2018imagenet}
Robert Geirhos, Patricia Rubisch, Claudio Michaelis, Matthias Bethge, Felix~A
  Wichmann, and Wieland Brendel.
\newblock Imagenet-trained cnns are biased towards texture; increasing shape
  bias improves accuracy and robustness.
\newblock {\em arXiv preprint arXiv:1811.12231}, 2018.

\bibitem{mummadi2021does}
Chaithanya~Kumar Mummadi, Ranjitha Subramaniam, Robin Hutmacher, Julien Vitay,
  Volker Fischer, and Jan~Hendrik Metzen.
\newblock Does enhanced shape bias improve neural network robustness to common
  corruptions?
\newblock In {\em International Conference on Learning Representations}, 2021.

\bibitem{szegedy2013intriguing}
Christian Szegedy, Wojciech Zaremba, Ilya Sutskever, Joan Bruna, Dumitru Erhan,
  Ian Goodfellow, and Rob Fergus.
\newblock Intriguing properties of neural networks.
\newblock {\em arXiv preprint arXiv:1312.6199}, 2013.

\bibitem{naseer2019cross}
Muzammal Naseer, Salman~H Khan, Harris Khan, Fahad~Shahbaz Khan, and Fatih
  Porikli.
\newblock Cross-domain transferability of adversarial perturbations.
\newblock {\em Advances in Neural Information Processing Systems}, 2019.

\bibitem{hendrycks2019benchmarking}
Dan Hendrycks and Thomas Dietterich.
\newblock Benchmarking neural network robustness to common corruptions and
  perturbations.
\newblock {\em arXiv preprint arXiv:1903.12261}, 2019.

\bibitem{li2017deeper}
Da~Li, Yongxin Yang, Yi-Zhe Song, and Timothy~M Hospedales.
\newblock Deeper, broader and artier domain generalization.
\newblock In {\em IEEE Conference on Computer Vision and Pattern Recognition},
  pages 5542--5550, 2017.

\bibitem{shao2021adversarial}
Rulin Shao, Zhouxing Shi, Jinfeng Yi, Pin-Yu Chen, and Cho-Jui Hsieh.
\newblock On the adversarial robustness of visual transformers.
\newblock {\em arXiv preprint arXiv:2103.15670}, 2021.

\bibitem{bhojanapalli2021understanding}
Srinadh Bhojanapalli, Ayan Chakrabarti, Daniel Glasner, Daliang Li, Thomas
  Unterthiner, and Andreas Veit.
\newblock Understanding robustness of transformers for image classification.
\newblock {\em arXiv preprint arXiv:2103.14586}, 2021.

\bibitem{brown2017adversarial}
Tom~B Brown, Dandelion Man{\'e}, Aurko Roy, Mart{\'\i}n Abadi, and Justin
  Gilmer.
\newblock Adversarial patch.
\newblock {\em arXiv preprint arXiv:1712.09665}, 2017.

\bibitem{paul2021vision}
Sayak Paul and Pin-Yu Chen.
\newblock Vision transformers are robust learners.
\newblock {\em arXiv preprint arXiv:2105.07581}, 2021.

\bibitem{brendel2019approximating}
Wieland Brendel and Matthias Bethge.
\newblock Approximating cnns with bag-of-local-features models works
  surprisingly well on imagenet.
\newblock {\em arXiv preprint arXiv:1904.00760}, 2019.

\bibitem{islam2021shape}
Md~Amirul Islam, Matthew Kowal, Patrick Esser, Sen Jia, Bjorn Ommer,
  Konstantinos~G Derpanis, and Neil Bruce.
\newblock Shape or texture: Understanding discriminative features in cnns.
\newblock {\em arXiv preprint arXiv:2101.11604}, 2021.

\bibitem{foster2011lower}
David~V Foster and Peter Grassberger.
\newblock Lower bounds on mutual information.
\newblock {\em Physical Review E}, 83(1):010101, 2011.

\bibitem{tuli2021convolutional}
Shikhar Tuli, Ishita Dasgupta, Erin Grant, and Thomas~L. Griffiths.
\newblock Are convolutional neural networks or transformers more like human
  vision?
\newblock {\em arXiv preprint arXiv:2105.07197}, 2021.

\bibitem{caron2021emerging}
Mathilde Caron, Hugo Touvron, Ishan Misra, Herv{\'e} J{\'e}gou, Julien Mairal,
  Piotr Bojanowski, and Armand Joulin.
\newblock Emerging properties in self-supervised vision transformers.
\newblock {\em arXiv preprint arXiv:2104.14294}, 2021.

\bibitem{zeiler2014visualizing}
Matthew~D Zeiler and Rob Fergus.
\newblock Visualizing and understanding convolutional networks.
\newblock In {\em European Conference on Computer Vision}, pages 818--833.
  Springer, 2014.

\bibitem{yang2013saliency}
Chuan Yang, Lihe Zhang, Huchuan Lu, Xiang Ruan, and Ming-Hsuan Yang.
\newblock Saliency detection via graph-based manifold ranking.
\newblock In {\em IEEE Conference on Computer Vision and Pattern Recognition},
  pages 3166--3173, 2013.

\bibitem{mao2021transformer}
Yuxin Mao, Jing Zhang, Zhexiong Wan, Yuchao Dai, Aixuan Li, Yunqiu Lv, Xinyu
  Tian, Deng-Ping Fan, and Nick Barnes.
\newblock Transformer transforms salient object detection and camouflaged
  object detection.
\newblock {\em arXiv preprint arXiv:2104.10127}, 2021.

\bibitem{vaswani2017attention}
Ashish Vaswani, Noam Shazeer, Niki Parmar, Jakob Uszkoreit, Llion Jones,
  Aidan~N Gomez, Lukasz Kaiser, and Illia Polosukhin.
\newblock Attention is all you need.
\newblock {\em arXiv preprint arXiv:1706.03762}, 2017.

\bibitem{he2016deep}
Kaiming He, Xiangyu Zhang, Shaoqing Ren, and Jian Sun.
\newblock Deep residual learning for image recognition.
\newblock In {\em IEEE Conference on Computer Vision and Pattern Recognition},
  pages 770--778, 2016.

\bibitem{forsyth2018probability}
David Forsyth.
\newblock {\em Probability and statistics for computer science}.
\newblock Springer, 2018.

\bibitem{hinton2015distilling}
Geoffrey Hinton, Oriol Vinyals, and Jeff Dean.
\newblock Distilling the knowledge in a neural network.
\newblock {\em arXiv preprint arXiv:1503.02531}, 2015.

\bibitem{hochreiter1997long}
Sepp Hochreiter and J{\"u}rgen Schmidhuber.
\newblock Long short-term memory.
\newblock {\em Neural Computation}, 9(8):1735--1780, 1997.

\bibitem{irie2019language}
Kazuki Irie, Albert Zeyer, Ralf Schl{\"u}ter, and Hermann Ney.
\newblock Language modeling with deep transformers.
\newblock {\em arXiv preprint arXiv:1905.04226}, 2019.

\bibitem{hendrycks2020augmix}
Dan Hendrycks, Norman Mu, Ekin~D. Cubuk, Barret Zoph, Justin Gilmer, and Balaji
  Lakshminarayanan.
\newblock {AugMix}: A simple data processing method to improve robustness and
  uncertainty.
\newblock {\em International Conference on Learning Representations}, 2020.

\bibitem{goodfellow2014explaining}
Ian~J Goodfellow, Jonathon Shlens, and Christian Szegedy.
\newblock Explaining and harnessing adversarial examples.
\newblock In {\em International Conference on Learning Representations}, 2014.

\bibitem{madry2017towards}
Aleksander Madry, Aleksandar Makelov, Ludwig Schmidt, Dimitris Tsipras, and
  Adrian Vladu.
\newblock Towards deep learning models resistant to adversarial attacks.
\newblock In {\em International Conference on Learning Representations}, 2018.

\bibitem{Razavian2014}
A.~Razavian, Hossein Azizpour, J.~Sullivan, and S.~Carlsson.
\newblock Cnn features off-the-shelf: An astounding baseline for recognition.
\newblock In {\em IEEE Conference on Computer Vision and Pattern Recognition
  Workshops}, pages 512--519, 2014.

\bibitem{cub}
P.~Welinder, S.~Branson, T.~Mita, C.~Wah, F.~Schroff, S.~Belongie, and
  P.~Perona.
\newblock {Caltech-UCSD Birds 200}.
\newblock Technical Report CNS-TR-2010-001, California Institute of Technology,
  2010.

\bibitem{Nilsback08}
M-E. Nilsback and A.~Zisserman.
\newblock Automated flower classification over a large number of classes.
\newblock In {\em Indian Conference on Computer Vision, Graphics and Image
  Processing}, 2008.

\bibitem{van2018inaturalist}
Grant Van~Horn, Oisin Mac~Aodha, Yang Song, Yin Cui, Chen Sun, Alex Shepard,
  Hartwig Adam, Pietro Perona, and Serge Belongie.
\newblock The inaturalist species classification and detection dataset.
\newblock In {\em IEEE Conference on Computer Vision and Pattern Recognition},
  pages 8769--8778, 2018.

\bibitem{aircraft}
S.~Maji, J.~Kannala, E.~Rahtu, M.~Blaschko, and A.~Vedaldi.
\newblock Fine-grained visual classification of aircraft.
\newblock Technical report, 2013.

\bibitem{dtd}
M.~Cimpoi, S.~Maji, I.~Kokkinos, S.~Mohamed, , and A.~Vedaldi.
\newblock Describing textures in the wild.
\newblock In {\em IEEE Conference on Computer Vision and Pattern Recognition},
  pages 3606--3613, 2014.

\bibitem{gtsrb}
Sebastian Houben, Johannes Stallkamp, Jan Salmen, Marc Schlipsing, and
  Christian Igel.
\newblock Detection of traffic signs in real-world images: The {G}erman
  {T}raffic {S}ign {D}etection {B}enchmark.
\newblock In {\em International Joint Conference on Neural Networks}, 2013.

\bibitem{fungi_dataset}
Brigit Schroeder and Yin Cui.
\newblock Fgvcx fungi classification challenge 2018.
\newblock In {\em \url{github.com/visipedia/fgvcx_fungi_comp, 2018}}.

\bibitem{zhou2017places}
Bolei Zhou, Agata Lapedriza, Aditya Khosla, Aude Oliva, and Antonio Torralba.
\newblock Places: A 10 million image database for scene recognition.
\newblock {\em IEEE Transactions on Pattern Analysis and Machine Intelligence},
  40(6):1452--1464, 2017.

\bibitem{meta_dataset}
Eleni Triantafillou, Tyler Zhu, Vincent Dumoulin, Pascal Lamblin, Kelvin Xu,
  Ross Goroshin, Carles Gelada, Kevin Swersky, Pierre{-}Antoine Manzagol, and
  Hugo Larochelle.
\newblock Meta-dataset: {A} dataset of datasets for learning to learn from few
  examples.
\newblock {\em http://arxiv.org/abs/1903.03096}, abs/1903.03096, 2019.

\bibitem{yonglong_rfs}
Yonglong Tian, Yue Wang, Dilip Krishnan, Joshua~B Tenenbaum, and Phillip Isola.
\newblock Rethinking few-shot image classification: a good embedding is all you
  need?
\newblock {\em arXiv preprint arXiv:2003.11539}, 2020.

\bibitem{yang2019fairer}
Kaiyu Yang, Klint Qinami, Li~Fei-Fei, Jia Deng, and Olga Russakovsky.
\newblock Towards fairer datasets: Filtering and balancing the distribution of
  the people subtree in the imagenet hierarchy.
\newblock In {\em ACM Conference on Fairness, Accountability, and
  Transparency}, pages 547--558, 2020.

\bibitem{yang2021study}
Kaiyu Yang, Jacqueline Yau, Li~Fei-Fei, Jia Deng, and Olga Russakovsky.
\newblock A study of face obfuscation in imagenet.
\newblock {\em arXiv preprint arXiv:2103.06191}, 2021.

\bibitem{liu2021Swin}
Ze~Liu, Yutong Lin, Yue Cao, Han Hu, Yixuan Wei, Zheng Zhang, Stephen Lin, and
  Baining Guo.
\newblock Swin transformer: Hierarchical vision transformer using shifted
  windows.
\newblock {\em IEEE International Conference on Computer Vision}, 2021.

\bibitem{radosavovic2020designing}
Ilija Radosavovic, Raj~Prateek Kosaraju, Ross Girshick, Kaiming He, and Piotr
  Doll{\'a}r.
\newblock Designing network design spaces.
\newblock In {\em IEEE Conference on Computer Vision and Pattern Recognition},
  2020.

\end{thebibliography}
}

\newpage
\appendix

\section{Random PatchDrop: Effect of Patch Size}
\label{sec:different_patch_Size_with_random_PatchDrop}

\begin{figure}[h!]
    \centering
    \begin{minipage}{\linewidth}
    \begin{minipage}{0.19\linewidth} \centering \tiny \textbf{4$\times$4 grid} \end{minipage}
    \begin{minipage}{0.19\linewidth} \centering \tiny \textbf{4$\times$8 grid} \end{minipage}
    \begin{minipage}{0.19\linewidth} \centering \tiny \textbf{7$\times$7 grid} \end{minipage}
    \begin{minipage}{0.19\linewidth} \centering \tiny \textbf{8$\times$8 grid} \end{minipage}
    \begin{minipage}{0.19\linewidth} \centering \tiny \textbf{16$\times$16 grid} \end{minipage}
    \end{minipage}
    \includegraphics[width=\linewidth]{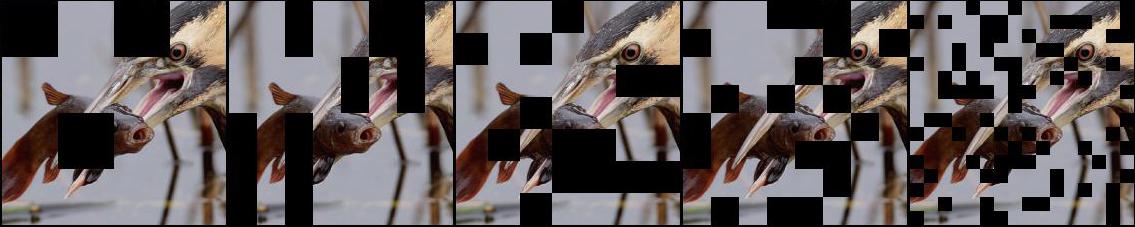}
    \caption{Visualization of varying grid sizes (resulting in different patch sizes) for PatchDrop experiments.}
    \label{app:vary_patchdrop_vis}
\end{figure}

\begin{figure}[h!]
    \centering
    \begin{minipage}{\linewidth}
    \includegraphics[width=0.33\linewidth]{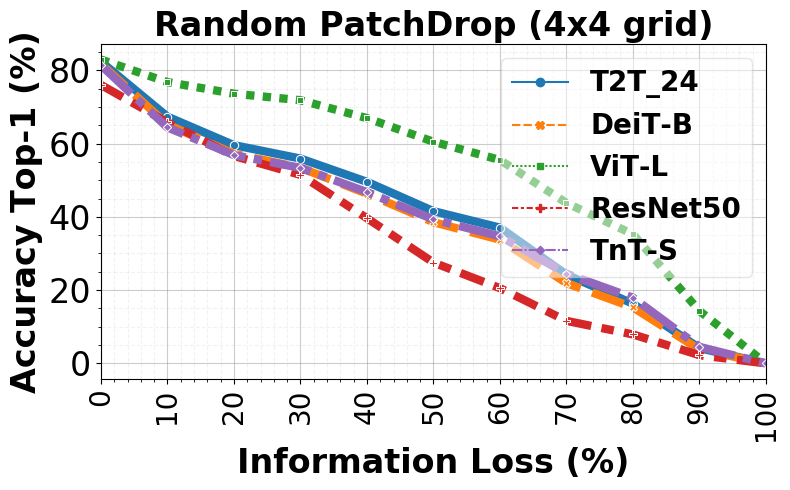}
    \includegraphics[width=0.33\linewidth]{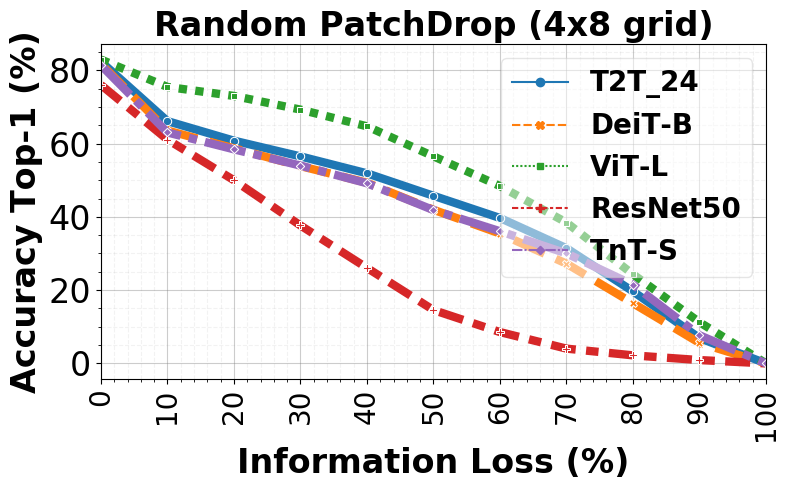}
    \includegraphics[width=0.33\linewidth]{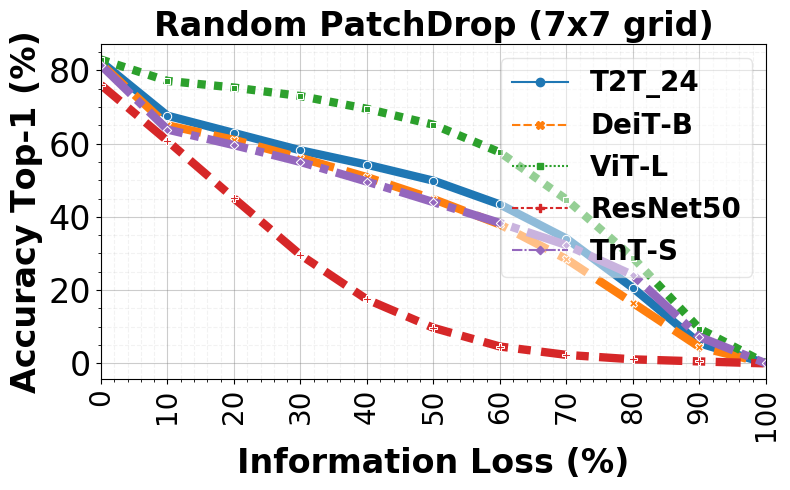}
    \end{minipage}
    \begin{minipage}{\linewidth}
    \begin{minipage}{0.66\linewidth}
    \includegraphics[width=0.5\linewidth]{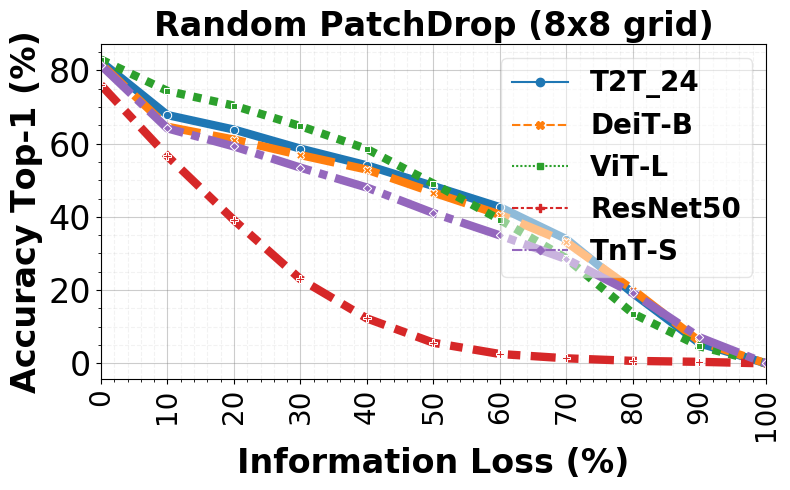}
    \includegraphics[width=0.5\linewidth]{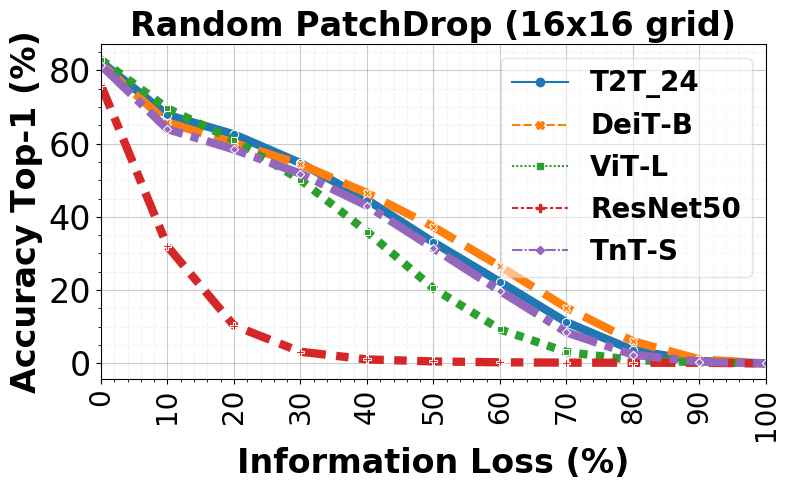}
    \end{minipage}
    \hspace{0.01\linewidth}
    \begin{minipage}{0.32\linewidth}
    \caption{Here, the patch sizes used for occlusion are different to the patch size used by ViT models (\emph{e.g.}, 16$\times$16 by default). Note that for larger patch sizes, the behaviour is closer to ResNet50, while for smaller patch sizes, ViT models generally perform better.}
    \label{app:vary_random_patch_drop}
    \end{minipage}
    \end{minipage}
\end{figure}

We extend our Random PatchDrop experiments to include varying patch sizes for the masking operation, as illustrated in Fig.~\ref{app:vary_patchdrop_vis}. The PatchDrop experiments in the main paper involved splitting the image into a 14$\times$14 grid (obtaining 196 patches of dimension 16$\times$16 pixels). Here, we split the image into different grid sizes and we define each experiment by the relevant grid size. Results for these experiments are presented in Fig.~\ref{app:vary_random_patch_drop}. All accuracy values are reported on the ImageNet val set. Since each grid size contains a different number of patches, we occlude a particular percentage and interpolate to the same scale in our accuracy plots for better comparison. 

We note that ViT models (that split an input image into a sequence of patches for processing) are significantly more robust to patch occlusion when dimensions of occluded patches are multiples of the model patch size (the grid size used is a factor of the original grid size). This is visible in the higher performance of ViT for the 7$\times$7 grid PatchDrop experiment (original uses 14$\times$14 grid).  At the same time, as large portions are occluded (\emph{e.g.}, with a 4$\times$4 spatial grid), the performance difference between ViT models and CNNs reduces considerably. We believe this to be the case since very large patch occlusions at high masking rates is likely to remove all visual cues relevant to a particular object category,  make it really challenging for both ViT and CNN models to make correct predictions.

More importantly, we note that the trends observed in Sec.~\ref{sec:patch_drop} about occlusions are reconfirmed from the varying grid-sizes experiment as well. We also note that some of these grid sizes (\emph{e.g.}, 8$\times$8) have no relation to the grid patterns used by the original ViT models (that split an image into a sequence of 14$\times$14 patches). This indicates that while these trends are more prominent for matching grid sizes (same as that of ViT models) and its factors, the observed trends are not arising solely due to the ViT models' grid operation. We note this behaviour is possible due to the dynamic receptive field of ViTs.

\subsection{Random PatchDrop with Offset}
\label{sec:patchdrop_offset}

\begin{figure}[t]
    \centering
    \begin{minipage}{0.38\linewidth}
    \includegraphics[width=\linewidth]{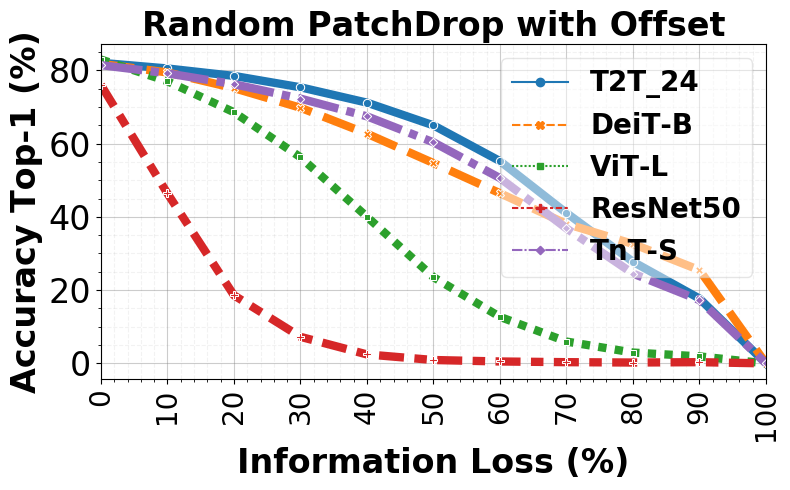}
    \end{minipage}
    \hspace{0.02\linewidth}
    \begin{minipage}{0.58\linewidth}
    \caption{We repeat our experiments in Sec.~\ref{sec:patch_drop} by adding an offset to the grid we use for masking patches. We aim to eliminate any biases due to any gird patterns that bear similarity with the kind of patches used by ViT models. To this end, in the PatchDrop experiments we remove alignment between our masks and ViT grid patterns. We note similar trends in this case as well, alongside a relative drop in ViT-L performance.}
    \label{app:offset_patchdrop}
    \end{minipage}
\end{figure}

We also explore how a spatial offset on our PatchDrop masks affects ViT models. This is aimed at eliminating the possible alignments between the intrinsic grid patterns of ViT models and our occlusion strategy, thereby avoiding any biasness in the evaluation setting towards a particular model family. The same masks are applied on the image, except with a small spatial offset to ensure that none of the masking patches align with any of the grid patterns used by ViT models in processing the input image. We replicate the same experiments as described in Sec.~\ref{sec:patch_drop} under this setting and present our results in Fig.~\ref{app:offset_patchdrop}. While in general we observe a similar trend between ViT models and the ResNet50 model, we note the significant drop of accuracy in ViT-L, in comparison to its performance under the no-offset setting. We present our potential reasons for this trend below.

ViT-L is a large-scale model containing over 300 million trainable parameters, while the other models contain significantly less parameters \emph{e.g.}, DeiT-B (86 million), T2T-24 (64 million), TnT-S (23 million), and ResNet50 (25 million). Furthermore, unlike ViT-L model, DeiT family and those building on it are trained with extensive data augmentation methods that ensure stable training of ViTs with small datasets. A similar relative drop of ViT-L performance is observed in the 16$\times$16 grid size experiment in Fig.~\ref{app:vary_random_patch_drop} as well. The anomalous behaviour of ViT-L in this setting is potentially owing to these differences. 

\section{Random PixelDrop}
\label{sec:random_pixeldrop}

\begin{figure}[b]
    \centering
    \begin{minipage}{\linewidth}
    \begin{minipage}{0.2\linewidth} \centering \tiny \textbf{Original} \end{minipage}
    \begin{minipage}{0.19\linewidth} \centering \tiny \textbf{10\%} \end{minipage}
    \begin{minipage}{0.2\linewidth} \centering \tiny \textbf{20\%} \end{minipage}
    \begin{minipage}{0.19\linewidth} \centering \tiny \textbf{30\%} \end{minipage}
    \begin{minipage}{0.2\linewidth} \centering \tiny \textbf{40\%} \end{minipage}
    \end{minipage}
    \includegraphics[width=\linewidth]{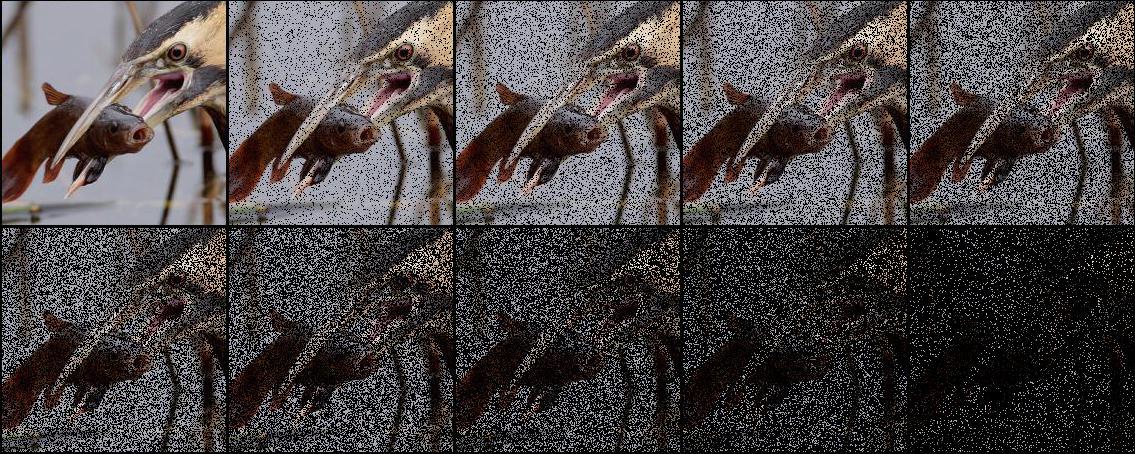}
    \begin{minipage}{\linewidth}
    \begin{minipage}{0.2\linewidth} \centering \tiny \textbf{50\%} \end{minipage}
    \begin{minipage}{0.19\linewidth} \centering \tiny \textbf{60\%} \end{minipage}
    \begin{minipage}{0.2\linewidth} \centering \tiny \textbf{70\%} \end{minipage}
    \begin{minipage}{0.19\linewidth} \centering \tiny \textbf{80\%} \end{minipage}
    \begin{minipage}{0.2\linewidth} \centering \tiny \textbf{90\%} \end{minipage}
    \end{minipage}
    
    \caption{Visualization of varying levels of PixelDrop (randomly masking pixels to study robustness against occlusions).}
    \label{app:vary_pixeldrop_vis}
\end{figure}

\begin{figure}[t]
    \centering
    \begin{minipage}{0.38\linewidth}
    \includegraphics[width=\linewidth]{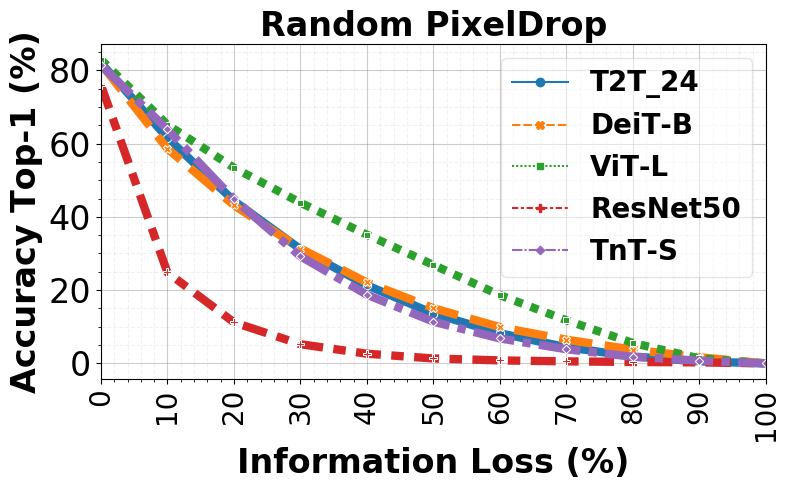}
    \caption{Random PixelDrop: we compare the performance of ViT models against a ResNet50 for our PixelDrop experiments illustrating how similar trends are exhibited.}
    \label{app:pixel_drop}
    \end{minipage}
    \hspace{0.02\linewidth}
    \begin{minipage}{0.58\linewidth}
    \includegraphics[width=\linewidth]{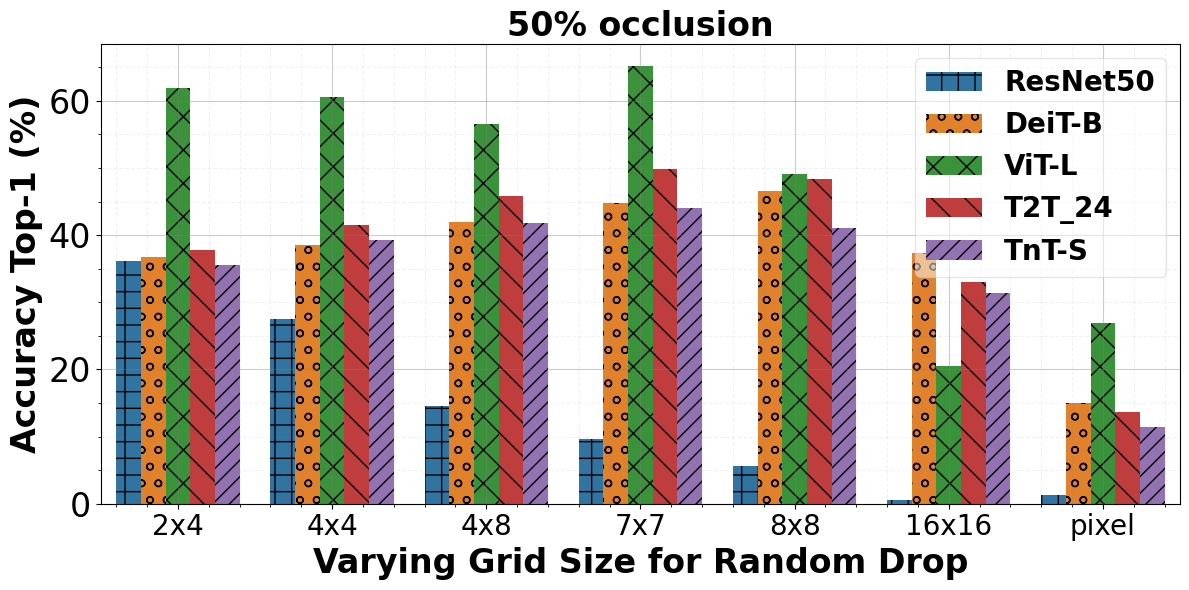}
    \caption{We compare the performance of models as we vary the grid size keeping the occlusion level constant at 50\% all the way until PixelDrop which we consider as PatchDrop with grid size equivalent to the image dimensions. While PixelDrop shows us similar trends as the occlusion level varies (Fig,~\ref{app:pixel_drop}), the general performance of models decreases.}
    \label{app:compare_grid_size}
    \end{minipage}
\end{figure}

A further step to observe the occlusion effect decoupled from the intrinsic grid operation of ViT models is to occlude at a pixel level. We generate pixel level masks of varying occlusion levels as illustrated in Fig.~\ref{app:vary_pixeldrop_vis}. Our evaluations on the ImageNet val. set presented in Fig.~\ref{app:pixel_drop} indicate the same trends between ViT models and CNNs that are observed earlier in Sec.~\ref{sec:patch_drop} and Appendix~\ref{sec:different_patch_Size_with_random_PatchDrop}. 

PixelDrop can be considered as a version of PatchDrop where we use a grid size equal to the image dimensions (setting patch size to 1$\times$1). Considering this, we compare how the performance of models varies as we approach PixelDrop from smaller grid sizes. This is illustrated in Fig.~\ref{app:compare_grid_size} where we evaluate models on the ImageNet val set at 50\% occlusion using PatchDrop with different grid sizes. We note that the overall performance of models drops for such fixed occlusion levels in the case of PixelDrop in comparison to the PatchDrop experiments.  

We also note how ViT-L displays significantly higher performance in comparison to the other models. This can be attributed to its much higher trainable parameter count as discussed in Sec.~\ref{sec:patchdrop_offset}. At the same time, ViT-L shows an anomalous drop in performance for the 16$\times$16 grid, quite similar to our observations in Fig.~\ref{app:offset_patchdrop}.

\section{Robustness to Feature Drop}
\label{sec:feature_drop}
In contrast to our previous experiments involving occlusion in the model input space, we now focus on occlusion within the model feature space. We achieve this by dropping portions of the intermediate representations inside the ViT model as opposed to dropping patches from the input image. For each transformer block (\emph{e.g.} for each of the 12 blocks in DeiT-B), we randomly mask (set to zero) a selected percentage of its input features. The effects of these ``feature drop'' experiments are studied in Table~\ref{app:lesion} by evaluating performance on the ImageNet val set. Performance is measured in the standard method (using the output of the final classifier head of the ViT model). We note that for small amounts of feature drop (25\% and 50\%), the models suffer relatively similar performance drops regardless of the individual block location. However, for larger amounts of feature drop, certain blocks emerge more important for each model. Furthermore, we notice a level of information redundancy within the blocks of larger models, as their performance drops are not significant even for considerable amounts of feature drop (e.g. ViT-L at 25\%).

\begin{SCtable}[\sidecaptionrelwidth][h!]
\vspace{0pt}
\centering
\begin{tabular}{l|c|c|c|c|c|c}
\toprule
\multirow{2}{*}{Block} & \multicolumn{3}{c|}{ViT-L} & \multicolumn{3}{c}{DeiT-B} \\ \cmidrule{2-7} 
                       & 25\%    & 50\%    & 75\%   & 25\%    & 50\%    & 75\%    \\ \midrule
Block 1                & 75.72   & 67.62   & 25.99  & 57.36   & 38.51   & 15.17   \\ 
Block 3                & 74.74   & 66.86   & 28.89  & 48.46   & 32.61   & 11.60   \\ 
Block 5                & 73.32   & 60.56   & 29.69  & 54.67   & 40.70   & 14.10   \\ 
Block 7                & 75.56   & 69.65   & 53.42  & 55.44   & 43.90   & 24.10   \\ 
Block 9                & 76.16   & 70.59   & 42.63  & 50.54   & 28.43   & 18.21   \\ 
Block 11               & 76.28   & 66.15   & 28.95  & 61.97   & 35.10   & 10.94   \\ \bottomrule
\end{tabular}
\vspace{0.5em}
\caption{Lesion Study: we drop a percentage of features input to each block of selected ViT models and evaluate their performance in terms of Top-1 accuracy (\%) on ImageNet val set. ViT-L shows significant robustness against such feature drop even up to the 25\% mark hinting towards information redundancy within the model.}
\label{app:lesion}
\end{SCtable}

\begin{SCtable}[\sidecaptionrelwidth][h!]
\setlength{\tabcolsep}{11pt}
\centering
\begin{tabular}{c|c|c|c}
\toprule
Block & 25\%  & 50\%  & 75\%  \\ \midrule
1     & 0.14  & 0.09  & 0.05  \\ 
2     & 45.09 & 4.91  & 0.23  \\ 
3     & 69.19 & 28.35 & 0.52  \\ 
4     & 73.95 & 64.12 & 18.95 \\ 
5     & 75.74 & 75.21 & 73.57 \\ \bottomrule
\end{tabular}
\caption{ResNet50 Lesion Study: we perform feature drop on the intermediate feature maps input to each of the four residual blocks (layers 1-4) and the feature map prior to the final average pooling operation (layer 5). We evaluate Top-1 accuracy (\%) on the ImageNet val. set for 25\%, 50\%, and 75\% feature drop applied to each layer.}
\label{resnet_lesion}
\end{SCtable}

In Table~\ref{resnet_lesion}, we conduct the same feature drop experiments for a ResNet50 model. We note that the ResNet50 architecture is entirely different to that of ViT models; hence comparison of these values will give little meaning. In the case of ResNet50, we observe how feature drop in the earlier layers leads to significant drops in performance unlike in ViT models. Also, feature drop in the last layer shows almost negligible drops in performance, which may be due to the average pooling operation which immediately processes those features. In the case of the ViT models compared, the patch tokens in the last layer are not used for a final prediction, so applying feature drop on them has no effect on the performance.  

\section{Robustness to Occlusions: More Analysis}
\label{sec:occultions_swin_regnety}

\begin{figure}[hb!]
    \centering
    \begin{subfigure}[b]{0.32\linewidth}        
        \centering
        \includegraphics[width=\linewidth]{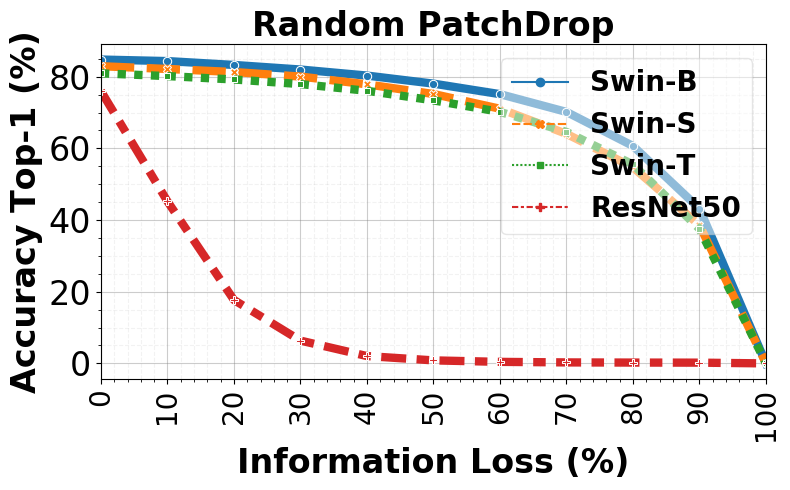}
    \end{subfigure}
    \begin{subfigure}[b]{0.32\linewidth}        
        \centering
        \includegraphics[width=\linewidth]{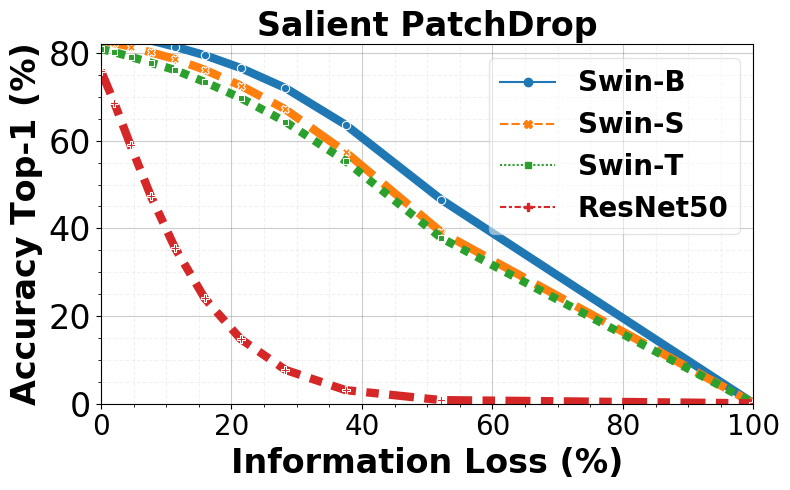}
    \end{subfigure}
    \begin{subfigure}[b]{0.32\linewidth}        
        \centering
        \includegraphics[width=\linewidth]{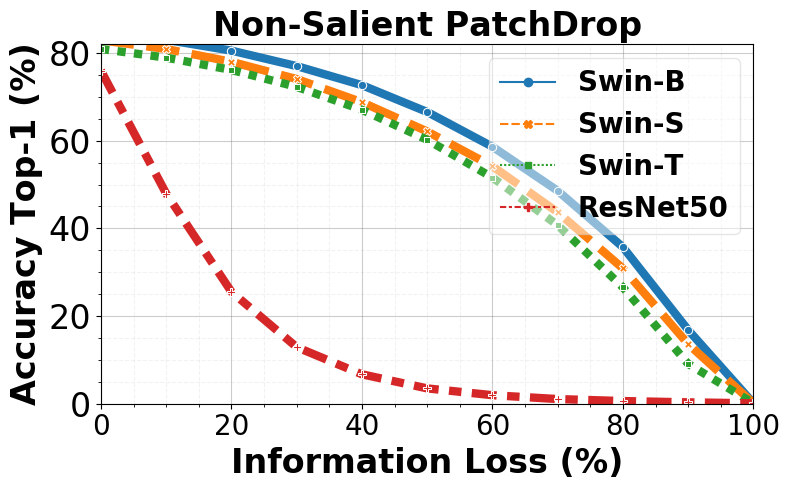}
    \end{subfigure}

     \caption{\small Robustness against object occlusion in images is studied under three PatchDrop settings (see Sec~\ref{sec:patch_drop}). We compare the Swin model family against ResNet50 exhibiting their superior robustness to object occlusion. These results show that ViT architectures that does not depend on using explicit class token like Swin transformer \cite{liu2021Swin} are  robust against information loss as well.}
     \label{fig:swin_transformer}
\end{figure}

In our experimental settings, we used ViTs with class tokens that interact with patch tokens throughout the network and are subsequently used for classification. However,  not all ViT designs use a class token e.g., Swin Transformer \cite{liu2021Swin} uses an average of all tokens. To this end, we conduct experiments (Fig.~\ref{fig:swin_transformer}) using three variants of the recent Swin Transformer \cite{liu2021Swin} against our proposed occlusions. 

\subsection{Swin Transformer \cite{liu2021Swin}}

\subsection{RegNetY \cite{radosavovic2020designing}}
\begin{figure}[hb!]
    \centering
    \begin{subfigure}[b]{0.32\linewidth}        
        \centering
        \includegraphics[width=\linewidth]{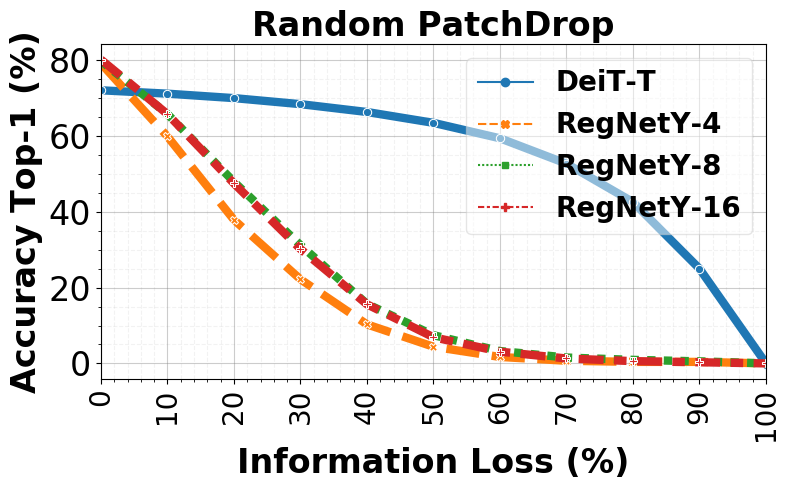}
    \end{subfigure}
    \begin{subfigure}[b]{0.32\linewidth}        
        \centering
        \includegraphics[width=\linewidth]{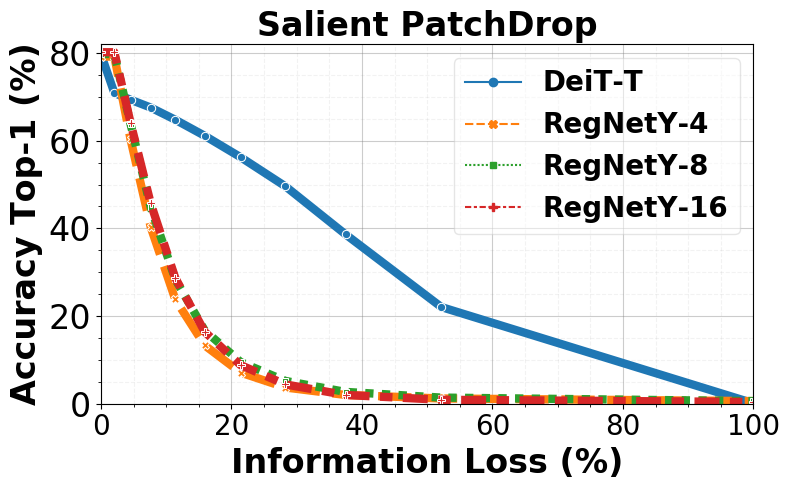}
    \end{subfigure}
    \begin{subfigure}[b]{0.32\linewidth}        
        \centering
        \includegraphics[width=\linewidth]{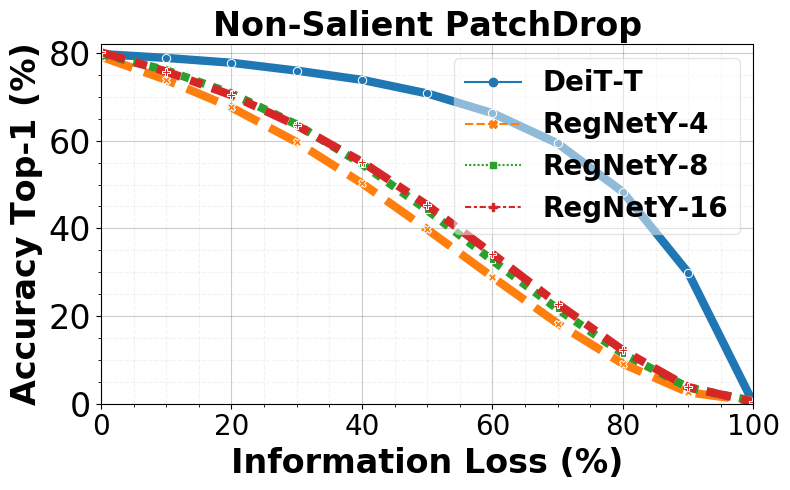}
    \end{subfigure}

     \caption{\small Robustness against object occlusion in images is studied under three PatchDrop settings (see Sec~\ref{sec:patch_drop}). We study the robustness of stronger baseline CNN model, RegNetY \cite{radosavovic2020designing} to occlusions, and identify that it overall behaves similar to other CNN models. Deit-T \cite{touvron2020deit}, a ViT with small number of parameters, performs significantly better than all the considered RegNetY variants.}
     \label{fig:regnety}
\end{figure}

Here, we evaluate three variants of RegNetY against our proposed occlusions (Fig.~\ref{fig:regnety}). RegNetY \cite{radosavovic2020designing} shows relatively higher robustness when compared to ResNet50, but overall behaves similar to other CNN models.

\newpage
\section{Behaviour of Shape Biased Models}
\label{sec:shape_biased}

In this section, we study the effect of our PatchDrop (Sec.~\ref{sec:patch_drop}) and permutation invariance (Sec.~\ref{sec:shuffle}) experiments on our models trained on Stylized ImageNet \cite{geirhos2018imagenet} (shape biased models). In comparison to a shape biased CNN model, the VIT models showcase  favorable robustness to occlusions presented in the form of PatchDrop. Note that ResNet50 (25 million) and DeiT-S (22 million) have similar trainable parameter counts, and therein are a better comparison. Furthermore, we note that in the case of ``random shuffle'' experiments, the ViT models display similar (or lower) permutation invariance in comparison to the CNN model. These results on random shuffle indicate that the lack of permutation invariance we identified within ViT models in Sec.~\ref{sec:shuffle} is somewhat overcome in our shape biased models.

\begin{figure}[h!]
    \centering
    \begin{minipage}{0.66\linewidth}
    \begin{minipage}{\linewidth}
    \includegraphics[width=0.5\linewidth]{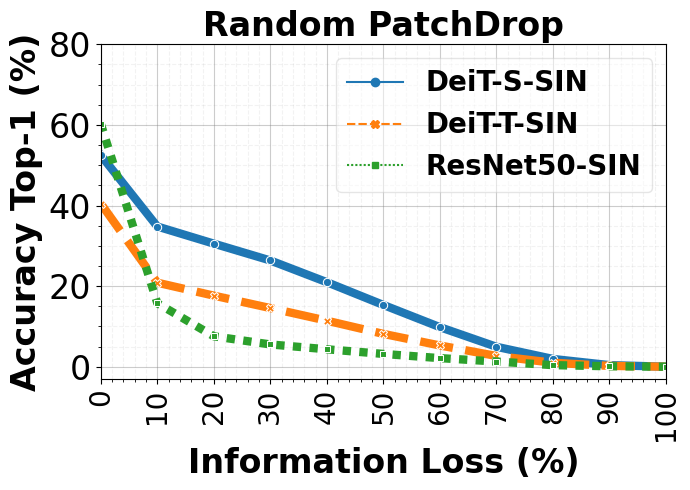}
    \includegraphics[width=0.5\linewidth]{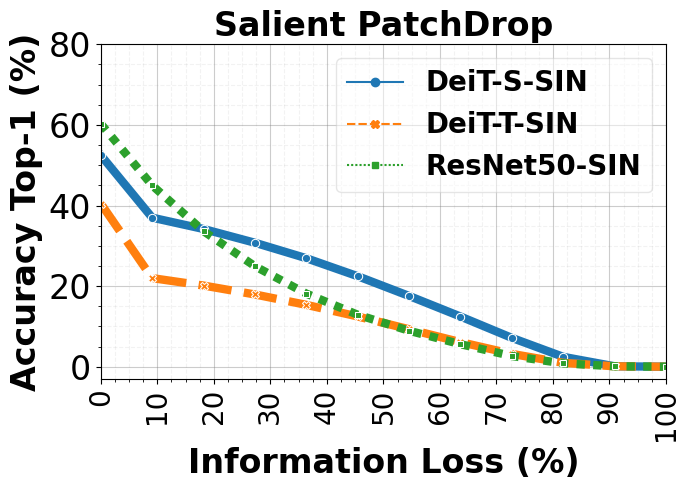}
    \end{minipage}
    \begin{minipage}{\linewidth}
    \includegraphics[width=0.5\linewidth]{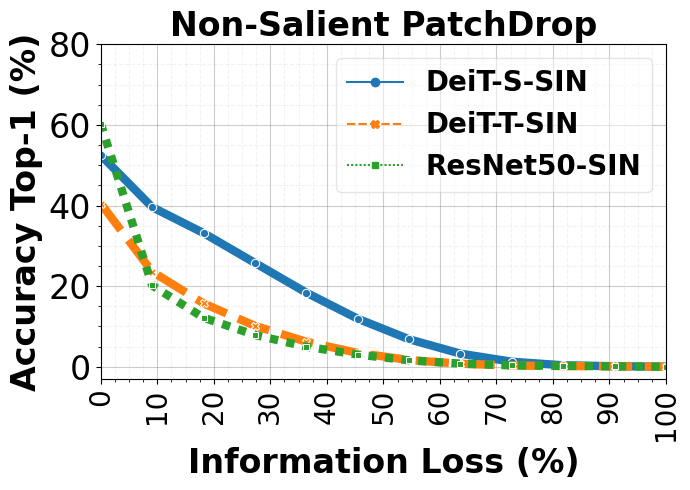}
    \includegraphics[width=0.5\linewidth]{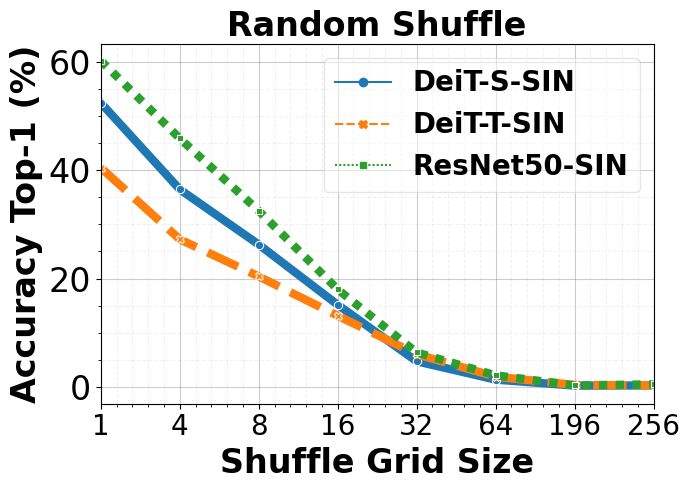}
    \end{minipage}
    \end{minipage}
    \hspace{0.02\linewidth}
    \begin{minipage}{0.30\linewidth}
    \caption{Shape biased models: We conduct the same PatchDrop and Random Shuffle experiments on DeiT models trained on Stylized ImageNet \cite{geirhos2018imagenet} and compare with a CNN trained on the same dataset. All results are calculated over the ImageNet val. set. We highlight the improved performance in the PatchDrop experiments for the DeiT models in comparsion to ResNet50. We also note how the DeiT models' performance drop with random shuffling is similar to that of the ResNet model.}
    \label{app:vary_random_patch}
    \end{minipage}
\end{figure}

\section{Dynamic Receptive field}
\label{sec:dynamic_receptive_field}

\begin{figure}[t!]
    \centering
    \begin{subfigure}[b]{0.32\linewidth}        
        \centering
        \includegraphics[width=\linewidth]{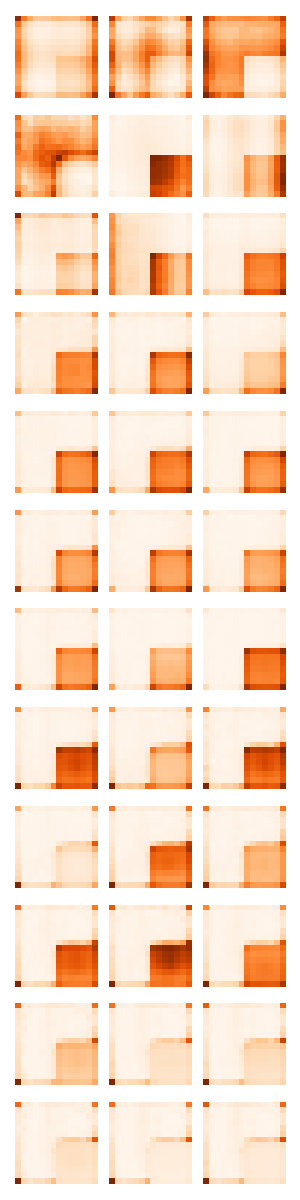}
        \caption{Bottom Right}
    \end{subfigure}
    \begin{subfigure}[b]{0.32\linewidth}        
        \centering
        \includegraphics[width=\linewidth]{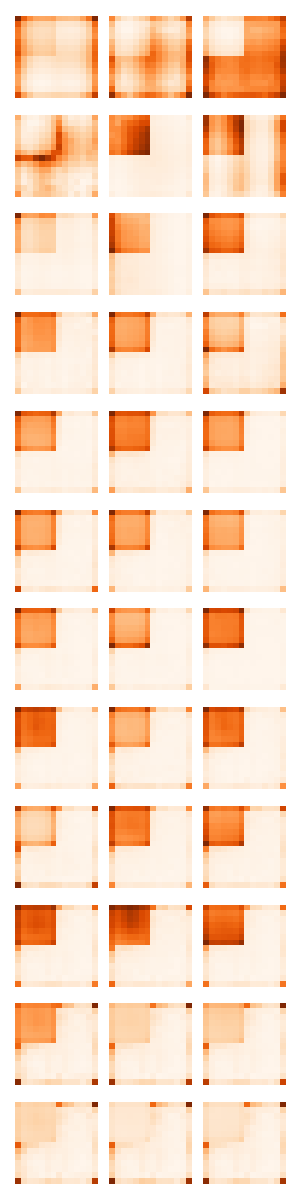}
        \caption{Upper Left}
    \end{subfigure}
    \begin{subfigure}[b]{0.32\linewidth}        
        \centering
        \includegraphics[width=\linewidth]{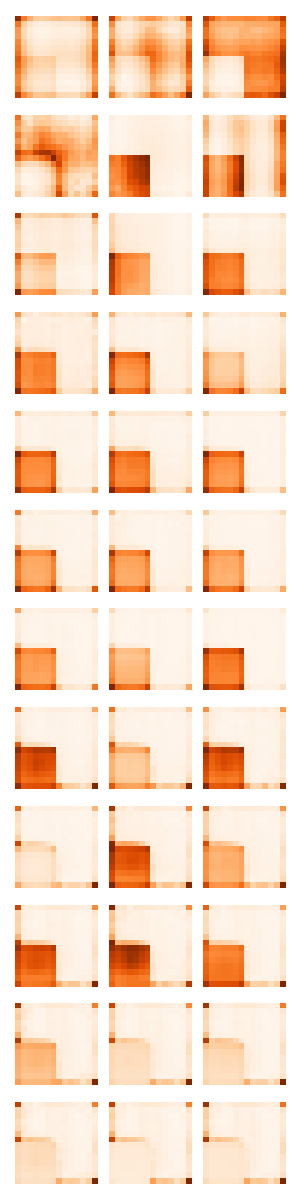}
        \caption{Bottom Left}
    \end{subfigure}

     \caption{\small Visualization depicting the change in attention, as the image is moved within the background. Attention maps (averaged over the entire ImageNet val. set) relevant to each head across all 12 layers of an ImageNet pre-trained DeiT-T (tiny) model \cite{touvron2020deit}. All images are rescaled to 128x128 and placed within black background. Observe how later layers clearly attend to non-occluded regions of images to make a decision, an evidence of the model's highly dynamic receptive field.}
     \label{fig:attention_shift_within_background}
\end{figure}

We further study the ViT behavior to focus on the informative signal regardless of its position. In our new experiment, during inference, we rescale the input image to 128x128 and place it within black background of size 224x224. In other words, rather than removing or shuffling image patches, we reflect all the image information into few patches. We then move the position of these patches to the upper/lower right and left corners of the background. On average, Deit-S shows 62.9\% top-1 classification accuracy and low variance (62.9$\pm$0.05). In contrast, ResNet50 achieves only 5.4\% top-1 average accuracy. These results suggest that ViTs can exploit discriminative information regardless of its position (Table~\ref{tbl:dynamic_receptive_field}). Figure \ref{fig:attention_shift_within_background} shows visualization depicting the change in attention, as the image is moved within the background. 

\begin{SCtable}[][h]\setlength{\tabcolsep}{3pt}
\centering
\scalebox{0.85}{
\small
\begin{tabular}{l c c c c }
\toprule
Models & top-right &	top-left&	bottom-right&	bottom-left  \\
\midrule
ResNet50&5.59&	5.71&	4.86&	5.30\\
\midrule
DeiT-T& 51.21&	51.38&	50.61&	50.70\\ 
\midrule
DeiT-S&63.14&	63.01&	62.62&	62.79\\  
\midrule
DeiT-B&\textbf{69.37}&\textbf{	69.29}&	\textbf{69.18}&\textbf{	69.20}\\ 
\bottomrule
\end{tabular}
}
\vspace{0.5em}
\caption{We rescale the input image to 128x128 and place it within the upper/lower right and left corners of the background of size 224x224. ViTs can exploit discriminative information regardless of its position as compared to ResNet50. Top-1 (\%) accuracy on ImageNet val. set is reported.}
\label{tbl:dynamic_receptive_field}
\end{SCtable}

\newpage
\section{Additional Qualitative Results}
\label{sec:supplementary_qualitative}
\begin{figure}[h]
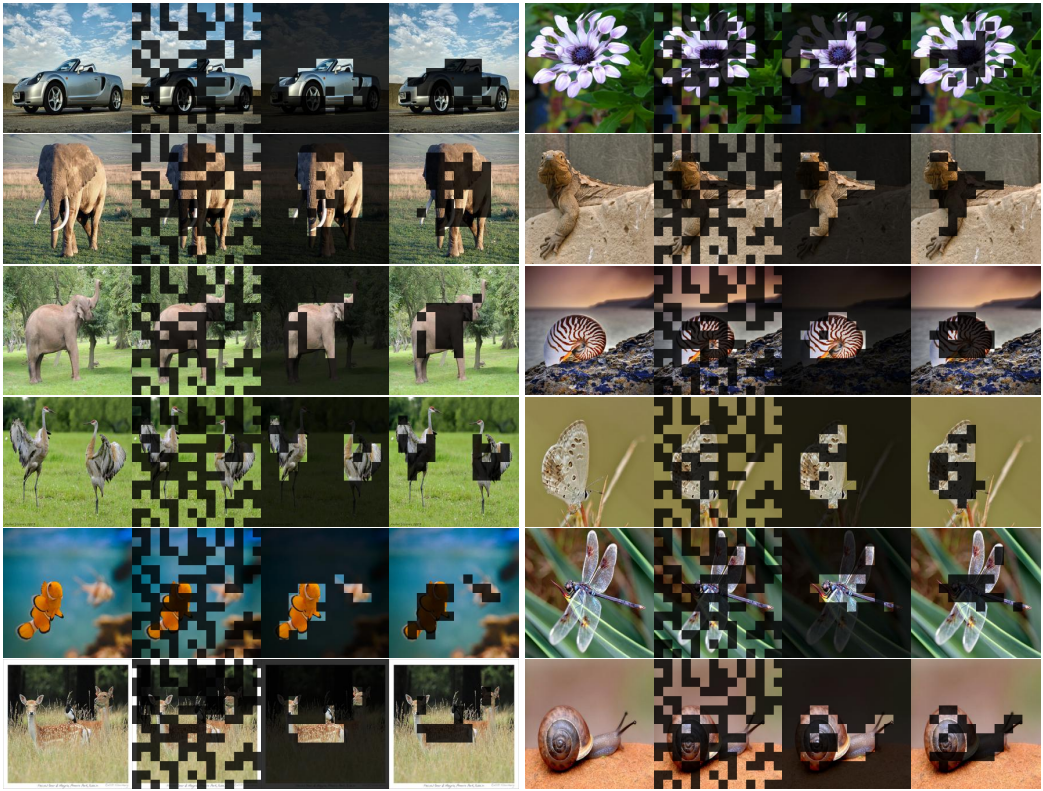

\centering
\foreach\x in {00004868,00008962,00008028,00036971,00035137,00037009}{%
    \begin{subfigure}[b]{0.49\linewidth}
        \includegraphics[width=0.25\linewidth]{figures/appendix/patch_drop/origILSVRC2012_val_\x.JPEG}%
        \includegraphics[width=0.25\linewidth]{figures/appendix/patch_drop/randomILSVRC2012_val_\x.JPEG}%
        \includegraphics[width=0.25\linewidth]{figures/appendix/patch_drop/background_removedILSVRC2012_val_\x.JPEG}%
        \includegraphics[width=0.25\linewidth]{figures/appendix/patch_drop/foreground_removedILSVRC2012_val_\x.JPEG}%
    \end{subfigure}
}%
\foreach\x in {00014376,00043407,00011419,00021739,00041135,00012418}{%
    \begin{subfigure}[b]{0.49\linewidth}
        \includegraphics[width=0.25\linewidth]{figures/appendix/patch_drop/origILSVRC2012_val_\x.JPEG}%
        \includegraphics[width=0.25\linewidth]{figures/appendix/patch_drop/randomILSVRC2012_val_\x.JPEG}%
        \includegraphics[width=0.25\linewidth]{figures/appendix/patch_drop/background_removedILSVRC2012_val_\x.JPEG}%
        \includegraphics[width=0.25\linewidth]{figures/appendix/patch_drop/foreground_removedILSVRC2012_val_\x.JPEG}%
    \end{subfigure}
}%
\caption{Visualizations for our three PatchDrop occlusion strategies: original, random (50\% w.r.t the image), non-salient (50\% w.r.t the forground predicted by DINO), and salient (50\% of the backgrond as predicted by DINO) PatchDrop (shown from \emph{left} to \emph{right}). DeiT-B model achieves accuracies of 81.7\%, 75.5\%, 68.1\%, and 71.3\% across the ImageNet val. set for each level of occlusion illustrated from \emph{left} to \emph{right}, respectively.}
\label{app:patchdrop_vis}
\end{figure}

\begin{figure}[h]
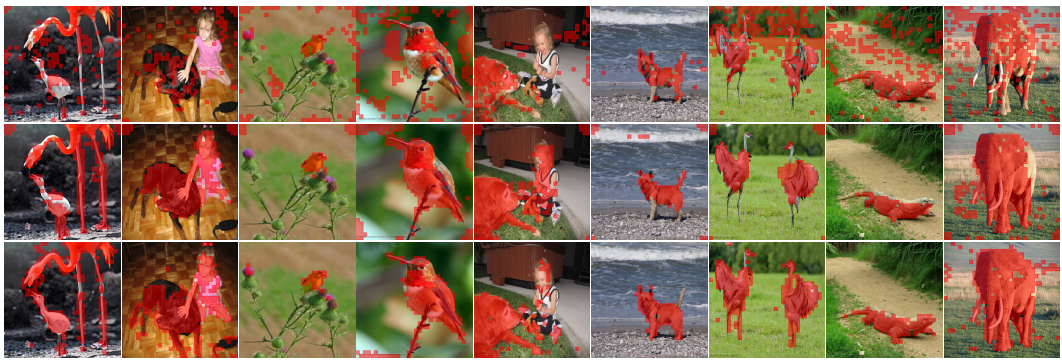

\centering
\begin{subfigure}[b]{\linewidth}        
\foreach\x in {1,...,9}{%
        \centering
        \includegraphics[width=0.11\linewidth]{figures/appendix/segmentation/nat\x.jpeg}
        \hspace{-0.01\linewidth}
}%
\end{subfigure}
\begin{subfigure}[b]{\linewidth}        
\foreach\x in {1,...,9}{%
        \centering
        \includegraphics[width=0.11\linewidth]{figures/appendix/segmentation/sin\x.jpeg}
        \hspace{-0.01\linewidth}
}%
\end{subfigure}
\begin{subfigure}[b]{\linewidth}        
\foreach\x in {1,...,9}{%
        \centering
        \includegraphics[width=0.11\linewidth]{figures/appendix/segmentation/dist\x.jpeg}
        \hspace{-0.01\linewidth}
}%
\end{subfigure}
\caption{Automatic segmentation of images using class-token attention for a DeiT-S model. Original, SIN trained, and SIN distilled model outputs are illustrated from \emph{top} to \emph{bottom}, respectively.}
\label{app:segmentation}
\end{figure}

\begin{figure}
\centering
\foreach\x in {0,1,3,4,5,6,7,8,9,10,11,12,14,15}{%
    \includegraphics[width=\linewidth]{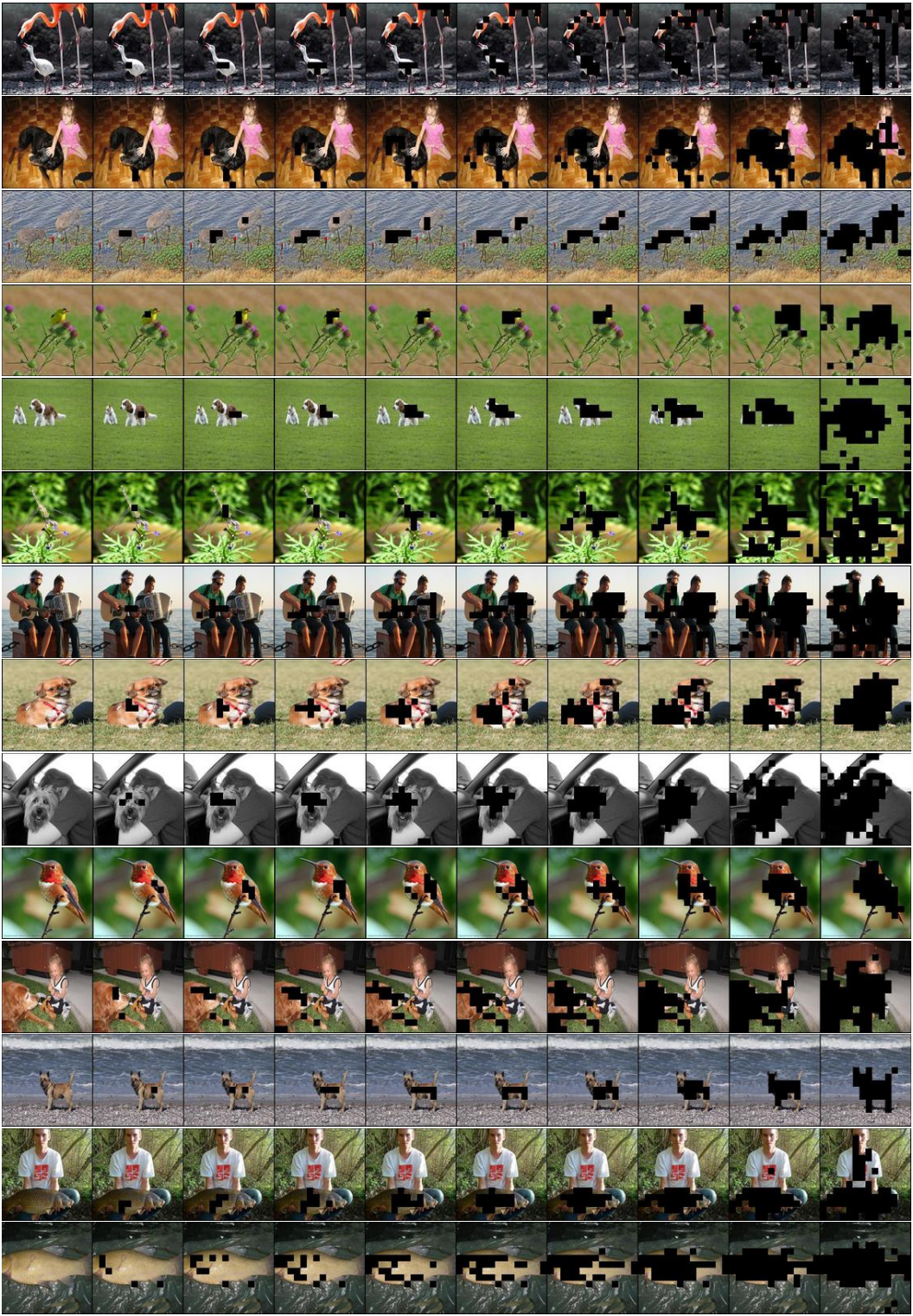}
}%
\caption{The variation (level increasing from \emph{left} to \emph{right}) of Salient PatchDrop on different images.}
\label{app:patchdrop_vary}
\end{figure}

\begin{figure}
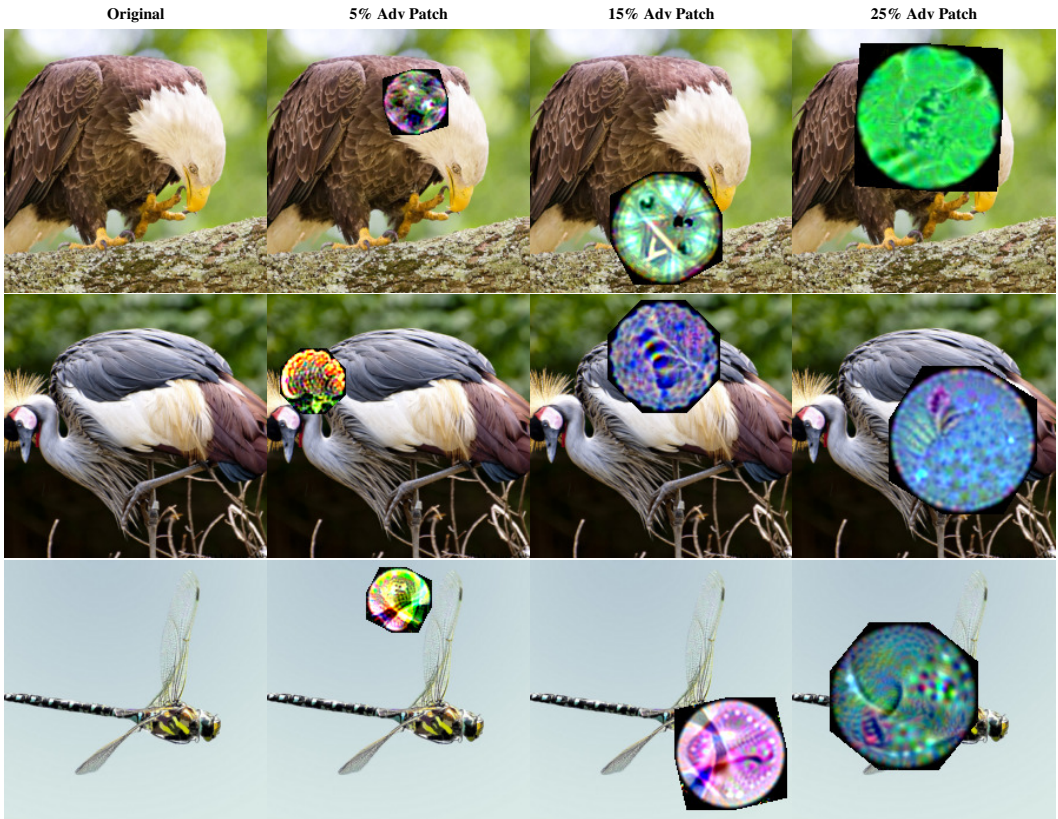

    \begin{subfigure}[b]{0.25\linewidth}%
        \caption*{\tiny\textbf{Original}}
    \end{subfigure}%
    \begin{subfigure}[b]{0.25\linewidth}%
        \caption*{\tiny\textbf{5\% Adv Patch}}
    \end{subfigure}%
    \begin{subfigure}[b]{0.25\linewidth}%
        \caption*{\tiny\textbf{15\% Adv Patch}}
    \end{subfigure}%
    \begin{subfigure}[b]{0.25\linewidth}%
        \caption*{\tiny\textbf{25\% Adv Patch}}
    \end{subfigure}%
    \vspace{-0.5em}
\foreach\y in {deit_tiny_patch16_224, deit_base_patch16_224, T2t_vit_24}{%
\begin{subfigure}[b]{\linewidth}
    \includegraphics[width=0.25\linewidth]{figures/appendix/adv/img_\y.jpeg}%
    \foreach\x in {0.05,0.15,0.25}{%
    \includegraphics[width=0.25\linewidth]{figures/appendix/adv/adv_\y_\x.jpeg}%
}%
\end{subfigure}
}%
\caption{Adversarial patches (universal and untargeted) optimized to fool DeiT-T, DeiT-B, and T2T-24 models from \emph{top} to \emph{bottom}. These ViT models are more robust to such adversarial patterns than CNN (e.g., ResNet50). }
\label{app:adv_patch}
\end{figure}

Here, we show some qualitative results, \emph{e.g.,} 
Fig. \ref{app:patchdrop_vis} show the examples of our occlusion (random, foreground, and background) method. The performance of our shape models to segment the salient image is shown in Fig. \ref{app:segmentation}.  We show different variation levels  of Salient PatchDrop on different images in Fig. \ref{app:patchdrop_vary}. Finally, we show adversarial patches optimized to fool different ViT models (Fig. \ref{app:adv_patch}).
\begin{figure}
\centering

\begin{subfigure}[b]{\linewidth}
    \begin{subfigure}[b]{0.25\linewidth}%
        \caption*{\tiny\textbf{Original}}
    \end{subfigure}%
    \begin{subfigure}[b]{0.25\linewidth}%
        \caption*{\tiny\textbf{5\% Adv Patch}}
    \end{subfigure}%
    \begin{subfigure}[b]{0.25\linewidth}%
        \caption*{\tiny\textbf{15\% Adv Patch}}
    \end{subfigure}%
    \begin{subfigure}[b]{0.25\linewidth}%
        \caption*{\tiny\textbf{25\% Adv Patch}}
    \end{subfigure}%
    \vspace{-0.5em}
    \includegraphics[width=0.25\linewidth]{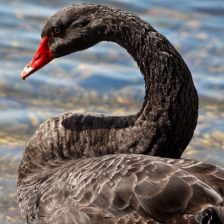}%
    \foreach\x in {0.05,0.15,0.25}{%
        \begin{subfigure}[b]{0.25\linewidth}%
        \includegraphics[width=\linewidth]{figures/appendix/adv/adv_deit_small_patch16_224_\x.jpeg}%
    \end{subfigure}%
}%
\end{subfigure}
\begin{subfigure}[b]{\linewidth}
    \includegraphics[width=0.25\linewidth]{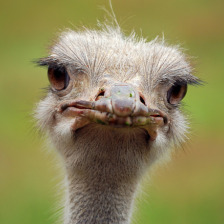}%
    \foreach\x in {0.05,0.15,0.25}{%
    \includegraphics[width=0.25\linewidth]{figures/appendix/adv/adv_deit_small_patch16_224_\x_SIN.jpeg}%
}%
\end{subfigure}
\caption{Adversarial patch (universal and untargeted) visualizations. \emph{Top} row shows adversarial patches optimized to fool DeiT-S trained on ImageNet, while \emph{bottom} row shows patches for DeiT-S-SIN. DeiT-S performs significantly better than DeiT-S-SIN. On the other hand, DeiT-SIN has higher shape-bias than DeiT-S.}
\label{app:adv_patch_sin}
\end{figure}

\end{document}